\documentclass[sigconf]{acmart}

\usepackage{booktabs} % For formal tables
\usepackage{times,amsmath,epsfig}
\usepackage{algorithmic}
\usepackage{algorithm}
\usepackage{multirow}
\usepackage{amssymb}
\usepackage{hyperref}
\usepackage{booktabs}
\usepackage{epstopdf}
\usepackage{graphicx}
\usepackage{subfigure}
\usepackage{color}

\newcommand{\eat}[1]{}

\eat{% DOI
\acmDOI{10.475/123_4}

\acmISBN{123-4567-24-567/08/06}

\acmConference[ACM Multimedia]{ACM Woodstock conference}{July 1997}{El
	Paso, Texas USA} 
\acmYear{2017}
\copyrightyear{2016}

\acmPrice{15.00}}

\begin{document}

	\title{Binary Generative Adversarial Networks for Image Retrieval}
	%\author{Paper ID: 2017}
	
	\author{Jingkuan Song}

\begin{abstract}
\eat{Generative Adversarial Networks (GAN) is able to generate synthetic data similar to training data from a latent space.
In this work, we propose a binary generative adversarial networks (BGAN), which restricts the latent space to be binary and conditioned on the features of training points, so that the latent space can be taken as a compact compression for the training data points.
generate plausible images, it must also learn to align them with the conditioning information,
Specifically, the latent space is generated by applying multi-layer perception (MLP) and a binary layer to a training image. Then, we train a GAN network to simultaneously map from a latent space to an image, and discriminate the true instances from the synthetic images. Furthermore, to be able to represent the original image, the latent space should reconstruct the synthetic image similar to the original image.
			
Experimental results on three large-scale datasets (i.e., CIFAR-10, NUSWIDE, and Flickr) demonstrate the superior performance of our BGAN over existing hashing methods.}

The most striking successes in image retrieval using deep hashing have mostly involved discriminative models, which require labels. In this paper, we use binary generative adversarial networks (BGAN) to embed images to binary codes in an unsupervised way. By restricting the input noise variable of generative adversarial networks (GAN) to be binary and conditioned on the features of each input image, BGAN can simultaneously learn a binary representation per image, and generate an image plausibly similar to the original one. In the proposed framework, we address two main problems: 1) how to directly generate binary codes without relaxation? 2) how to equip the binary representation with the ability of accurate image retrieval? We resolve these problems by proposing new sign-activation strategy and a loss function steering the learning process, which consists of new models for adversarial loss, a content loss, and a neighborhood structure loss. Experimental results on standard datasets (CIFAR-10, NUSWIDE, and Flickr) demonstrate that our BGAN significantly outperforms existing hashing methods by up to 107\% in terms of~mAP (See Table \ref{tab.res.map.comp})\footnote{Our anonymous code is available at: \url{https://github.com/htconquer/BGAN}}.
	
%To address these issues, first, to solve the non-smooth sign activation, we propose two equivalent but smoothed activation functions, and design a learning strategy whose solution converges to the results of sign activation.
%Second, we propose a loss function which consists of an adversarial loss, a content loss, and a neighborhood structure loss. The adversarial loss pushes the generated image to the natural image manifold using a discriminator network that is trained to differentiate between the generated images and original images. The content loss enforces the generated image to be aligned with the conditioned original image, and a neighborhood structure loss is to exploit the data distributions of the images feature space.
	
\end{abstract}


\maketitle
\vspace{-0.39cm}
\section{Introduction}
\label{sec.intro}
With the rapidly increasing amount of images, similarity search in large image collections has been actively pursued in a number of domains, including computer vision, information retrieval and pattern recognition~\cite{Moraleda08,WangSSJ14}. However, exact nearest-neighbor (NN) search is often intractable because of the size of dataset and the high dimensionality of images. Instead, approximate nearest-neighbor (ANN) search is more practical and can achieve orders of magnitude in speed-up compared to exact NN search~\cite{Moraleda08,JegouDS11}.
	
Recently, learning-based hashing methods \cite{0002F13,WangSSJ14,IrieLWC14,LinSSHS14,HuJRCH14,WangSZ14} have become the mainstream for scalable image retrieval due to their compact binary representation and efficient Hamming distance calculation. Such approaches embed data points to compact binary codes through hash functions, which can be generally expressed as:
	\begin{equation}
	{\textbf{b}} = {\textbf{h}}\left( {\textbf{x}} \right) \in {\left\{ {0,1} \right\}^L}
	\end{equation}
	where ${\textbf{x}} \in \mathbb{R}^{M\times 1}$, ${\textbf{h}}(.)$ are the hash functions, and ${\textbf{b}}$ is a binary vector with code length $L$. 
	
Hashing methods can be generally categorized as being unsupervised or supervised. The unsupervised learning of a hash function is usually based on the criterion of preserving important properties of the training data points in the original space. Typical approaches target pairwise similarity preservation (i.e., the similarity/distance of binary codes should be consistent with that of the original data points) \cite{Weiss08,LiuHDL14,IrieLWC14}, multi-wise similarity preservation (i.e., the similarity orders over more than two items computed from the input space and the coding space should be preserved) \cite{NorouziF11,WangWYL13},  or implicit similarity preservation (i.e., pursuing effective space partitioning without explicitly evaluating the relation between the distances/similarities in the input and coding spaces), \cite{HeoLHCY15,JinHLZLCL13}. A fundamental limitation of a hashing method geared to preserve a particular image property is that its performance may degrade when it is is applied to a context where a different property is relevant.
	
Supervised hashing is designed to generate the binary codes based on predefined labels~\cite{LinSSHS14,GeH014,strecha2012ldahash}. For example, Strecha \textit{et al}. \cite{strecha2012ldahash} developed a supervised hashing which maximizes the between-class Hamming distance and minimizes the within-class Hamming distance. \cite{LinSSHS14,GeH014} proposed to learn the hash codes such to approximate the pairwise label similarity. Supervised hashing methods usually significantly outperform unsupervised methods. However, the information that can be used for supervision is also typically scarce.

	\eat{\begin{figure}[t]
			\centering
			\hspace{-1.7cm}
			\includegraphics[width=1.05\linewidth]{scene3.pdf}
			\hspace{-1.7cm}
			\caption{Reconstructed images using binary GAN. (\textcolor{blue}{Replace this image})}
			\label{fig.mapVSpe}
		\end{figure}}

More recently, deep learning has been introduced in the development of hashing algorithms~\cite{CNNH,guo2016hash,dai2016binary,lin2016learning,do2016learning,yang2017supervised,gu2016supervised}, leading to a new generation of \textit{deep hashing} algorithms. Due to powerful feature representation, remarkable image retrieval performance has been reported using the hashes obtained in this way. However, a number of open issues have still remain open. The most successful deep hashing methods are usually supervised and require labels. The labels are, however, scarce and subjective. Unsupervised approaches, on the other hand, cannot take full advantages of the current deep learning models, and thus yield unsatisfactory performance \cite{lin2016learning}. Another issue is a non-smooth sign-activation function used to generate the binary codes, which, despite several ideas proposed to tackle it~\cite{Li2015Feature,cao2017hashnet,do2016learning}, still makes the standard back-propagation infeasible. 
		
%I WOULD SUGGEST TO USE THIS IN THE RELATED WORK SECTION: Several ideas have been proposed to tackle this issue. For instance3, in \cite{cao2017hashnet}, Cao \textit{et. al.} proposed a tanh-like function to replace the sign activation and Li \textit{et. al.} introduced an intermediate continuous variable to approximate the binary code. In \cite{Li2015Feature,do2016learning}, they actually relaxed the binary constraints, and the binary codes were further generated by a sign function. And the work of \cite{cao2017hashnet} only considered a special case of solving the non-smooth sign function, instead of looking for a generalized solution.
		
To address the above issues, we propose an unsupervised hashing method that deploys a generative adversarial network (GAN) \cite{Reed2016Generative}. GAN has proven effective to generate synthetic data similar to the training data from a latent space. Therefore, if we restrict the input noise variable of generative adversarial networks (GAN) to be binary and conditioned on the features of each input image, we can learn a binary representation for each image and generate a plausibly similar image to the original one simultaneously. Feeding the generated images through a ``discriminator'' that verifies them with respect to the training images removes the need for supervision and the relevant hash can be learned in an unsupervised fashion. We refer to this proposed architecture as \textit{binary GAN} (BGAN). For the BGAN learning process, we design a novel loss function to equip the binary representation with the ability of accurate image retrieval, beyond vivid image generation. Furthermore, inspired by recent studies on continuation methods \cite{allgower2012numerical,cao2017hashnet}, we propose two equivalent realizations of the sign-activation function, and design an optimization strategy whose solution is equivalent to the non-smooth sign function. 
		
%The remainder of this paper is organized as follows. In Section 2, we review the existing work related to the problem of video hashing and position our proposed solution with respect to it. Section 3 presents our proposed BGAN image hashing method. The method is evaluated experimentally in Section 4. Section 5 concludes the papers by a discussion of the obtained image retrieval performance and pointers to future work on image hashing.  

\vspace{-0.2cm}	
\section{Related Work}
\subsection{Hashing}
Existing hashing methods can be generally divided into two categories: unsupervised, and supervised methods. For unsupervised methods \cite{Gong2013Iterative,Salakhutdinov2009Semantic,dai2016binary,lin2016learning,erin2015deep}, label information is not required in the learning process. For example, Lin \textit{et al.}~\cite{lin2016learning} proposed an unsupervised deep learning approach, DeepBit, imposing three criteria on binary codes (i.e., minimal loss quantization, evenly distributed codes and uncorrelated bits) to learn a compact binary descriptor for efficient visual object matching. The ITQ method proposed by Gong \textit{et al.}~\cite{Gong2013Iterative} maximizes the variance of each binary bit and minimizes the binarization loss to obtain a high performance for image retrieval. Liong \textit{et.al} \cite{erin2015deep} proposed to use a deep neural network to learn hash codes by optimizing for three objectives: (1) the loss between the real-valued feature descriptor and the learned binary codes is minimized, (2) binary codes distribute evenly on each bit and (3) different bits are as independent as possible.
		
Supervised hashing methods \cite{CNNH,guo2016hash,do2016learning,Li2015Feature, yang2017supervised,gu2016supervised} utilize the label information and can usually obtain better performance. Most of the methods apply pairwise label loss as criterion for optimization. An example is the deep supervised hashing method by Li \textit{et.al} \cite{Li2015Feature}, which can simultaneously learn features and hash codes. Another example is Supervised Recurrent Hashing (SRH) for creating video hashes, as proposed by Gu \textit{et. al.}~\cite{gu2016supervised}. Cao \textit{et.al} \cite{cao2017hashnet} proposed a continuous method to learn binary codes, which can avoid the relaxation of binary constraints~\cite{gu2016supervised} by first learning continuous representations and then thresholding them to get the hash codes. They also added weight to data for balancing similar and dissimilar pairs.

So far, only supervised hashing methods have been reasonably successful in benefiting from the effectiveness of deep learning. In this paper we address the challenge of enabling an unsupervised method to also take full advantage of this machine learning approach and generate reliable hash codes without the need for class labels.
	
We also address a specific issue related to the deployment of a deep learning approach in the hashing context, namely that a non-smooth sign-activation function used to generate the binary codes still makes the standard back-propagation infeasible. With our new proposed sign-activation strategy, we provide an effective alternative to earlier attempts to address this issue~\cite{Li2015Feature,cao2017hashnet,do2016learning}. For instance, in \cite{cao2017hashnet}, Cao \textit{et. al.} proposed a tanh-like function to replace the sign activation and Li \textit{et. al.} introduced an intermediate continuous variable to approximate the binary code~\cite{Li2015Feature,do2016learning}. Essentially, they relaxed the binary constraints and the binary codes were further generated by a sign function. Furthermore, the work of Cao \textit{et. al.}~\cite{cao2017hashnet} only considered a special case of solving the non-smooth sign function, instead of looking for a generalized solution.

\vspace{-0.3cm}		
\subsection{Generative Adversarial Nets}
Ian \textit{et. al.}~\cite{goodfellow2014generative} proposed Generative Adversarial Network (GAN), which has been widely applied in computer vision and natural language processing. In the nutshell, GAN is an architecture for playing a min-max game, where two competing processes aim at maximizing their individual, but mutually conflicting, objectives. GAN has shown its effectiveness in various application contexts and various variants of the original GAN concept have been developed \cite{radford2015unsupervised,larsen2015autoencoding,Chen2016InfoGAN}. Radford \textit{et. al.} \cite{radford2015unsupervised} proposed a method for unsupervised feature learning using deep convolutional generative adversarial networks (DCGAN), which was shown to be superior to other unsupervised algorithms. In \cite{larsen2015autoencoding}, Larsen \textit{et. al.} combined a variational autoencoder (VAE)~\cite{kingma2013auto} and GAN to use learned feature representations in the GAN discriminator as the basis for the VAE reconstruction objective. This approach was shown to outperform VAEs with element-wise similarity measures in terms of visual fidelity. In \cite{Chen2016InfoGAN}, Chen \textit{et. al.} proposed a model called Variational InfoGAN (ViGAN), which not only can generate new images based on visual descriptions, but can also retain the latent representation of an image and varying the visual description. 

Nevertheless, the learned representations serving as input into the GAN generator in the above methods are not binary. Using a binary image representation in combination with a GAN framework is a new challenge that we address in this paper by proposing \textit{binary generative adversarial networks} (BGAN).

\begin{figure*}[t]
	\centering
	\includegraphics[width=0.85\linewidth,height=6cm]{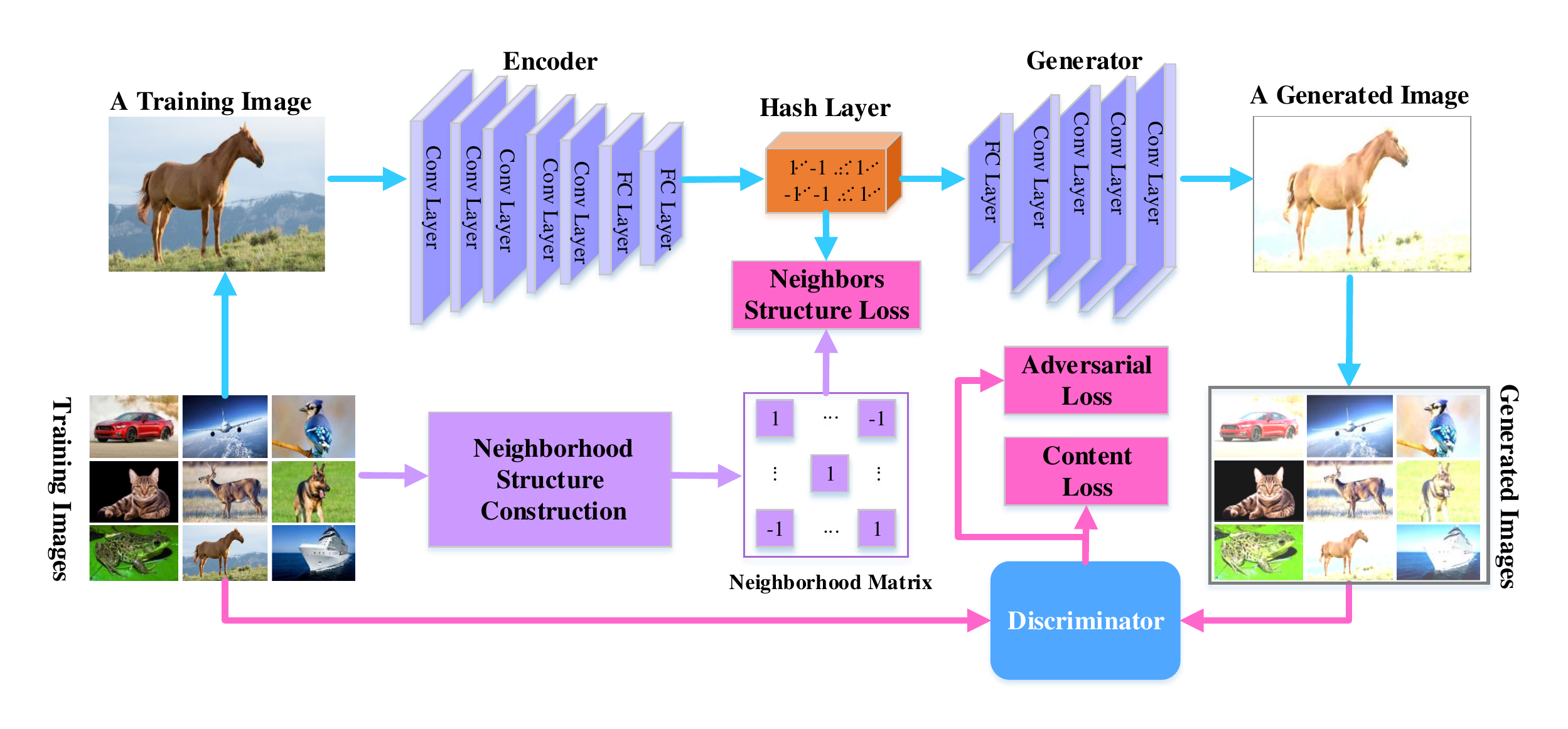}
	\caption{The proposed framework for BGAN, which is comprised of four key components: (1) an encoder, for learning image representations, (2) a hashing layer, for embedding the L-dimensional representation into L-bit binary hash code, (3) a decoder, to reconstruct the original images, and (4) a discriminator, for distinguishing real and reconstructed images. As a pre-processing step, we construct the neighborhood structure of the training images.}
	\label{fig.framework}
\end{figure*}

\vspace{-0.15cm}
\section{Binary Generative Adversarial Networks}
Given $N$ images, $\textbf{I} = \{\textbf{I}_i\}^N_{i=1}$ without labels, our goal is to learn their compact binary codes $\textbf{B}$ and reconstructed images $\textbf{I}^R$ for the original images such that: (a) the binary codes can reconstruct the image content, and (b) the binary codes could be computed directly without relaxation.

We illustrate our proposed BGAN architecture by the scheme in Fig.~\ref{fig.framework}. The scheme shows how hash codes are learned in an unsupervised fashion through the interplay between, on the one side, the image generation process (the generator module) taking the generated hash code (the hash layer) as input, and, on the other side, the verification process (the discriminator module) where the images from the generator are compared to the original training images. 
%The production of the hash codes takes as input the result of image encoding (the encoder module) and is steered by the information extracted from the visual neighbors of the target image by means of minimizing the \textit{neighborhood structure loss}. The verification in the discriminator, on the other hand, is steered through the optimization with respect to the \textit{adversarial loss} and \textit{content loss}. For training of the system, we first construct the neighborhood structure of images and then train the neural network underlying the encoder, generator and discriminator. The training of the system is converged when the generator manages to deceive the discriminator, that is, when the discriminator is not capable any more of differentiating between the original and reconstructed images. 
%In the remainder of this section, we describe in detail our realization of different components of the scheme and in particular our proposed models for the three above mentioned loss functions.
For training of the system, we first construct the neighborhood structure of images and then train the neural network underlying the encoder, generator and discriminator.
In the remainder of this section, we describe the process of constructing the neighborhood structure, the network architecture, our loss function and learning of parameters.

\eat{Given $N$ images, $\textbf{I} = \{\textbf{I}_i\}^N_{i=1}$ without labels, our goal is to learn their compact binary codes $\textbf{B}$ and reconstructed images $\textbf{I}^R$ for the original images such that: (a) the binary codes can reconstruct the image content, and (b) the binary codes could be computed directly without relaxation.

As shown in Fig.~\ref{fig.framework}, our BGAN consists of four key components: encoder, hashing, generator and discriminator. For training, we first construct the neighborhood structure of images and then train the network. For testing, we obtain the binary codes of an image by taking it as an input. In the remainder of this section, we describe the process of constructing the neighborhood structure, the network architecture, our loss function and the process of learning the parameters.}

\vspace{-0.14cm}
\subsection{Construction of Neighborhood Structure}
In our unsupervised approach, we propose to exploit the neighborhood structure of the images in a feature space as information source steering the process of hash learning. Specifically, we propose a method based on the K-Nearest Neighbor (KNN) concept to create a neighborhood matrix $\textbf{S}$. Based on \cite{He2015}, we extract 2,048-dimensional features from the pool5-layer. This results in the set $\textbf{X} = \{{\textbf{x}_{i}}\}^{N}_{i=1}$ where $\textbf{x}_{i}$ is the feature vector of image $\textbf{I}_i$.

For the representation of the neighboring structure, our task is to construct a matrix $\textbf{\textit{S}}=\{s_{ij}\}^N_{i,j=1}$, whose elements indicate the similarity ($s_{ij}=1$) or dissimilarity ($s_{ij} = -1$) of any two images $i$ and $j$ in terms of their features $\textbf{x}_{i}$ and $\textbf{x}_{j}$.

\begin{figure}[h]
	\centering
	\small
	\includegraphics[width=0.8\linewidth,height=3.8cm]{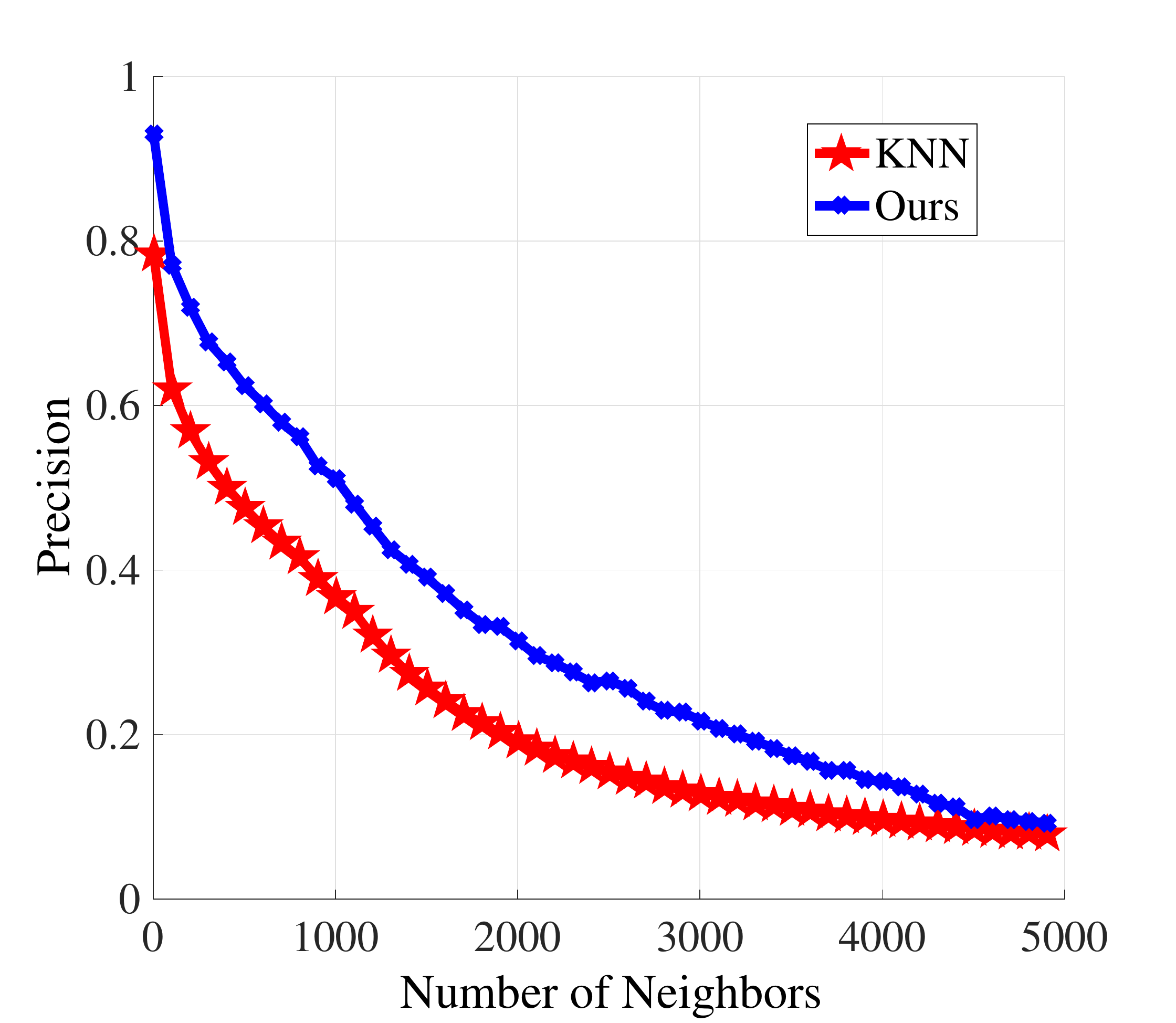}
	\caption{Precision of constructed labels on cifar-10 dataset with different K, and different methods.}
	\label{fig1}
\end{figure}

We compare images using cosine similarity of the feature vectors. For each image, we select $K1$ images with the highest cosine similarity as its neighbors. Then we can construct an initial similarity matrix $\textbf{S}_1$:
\begin{eqnarray}
{{\left(S_1\right)}_{ij}} = \left\{ {\begin{array}{*{20}{l}}
	{1,~\textrm{if}~\textbf{x}_{j}~\textrm{is K1-NN of}~\textbf{x}_i}\\
	{-1,~\textrm{otherwise}}
	\end{array}} \right.
\end{eqnarray}

\begin{algorithm}[t]
	\begin{algorithmic}[1]
		\renewcommand{\algorithmicrequire}{\textbf{Input:}}
		\renewcommand{\algorithmicensure}{\textbf{Output:}}
		\REQUIRE Images $\textbf{X}  = \{ {\textbf{x}_i}\} _{i = 1}^N$, the number of neighbors K1, the number of neighbors K2 for the neighbors expansion;
		\ENSURE Neighborhood matrix $\textbf{S} = \{ {s_{ij}}\}$;
		\STATE	{First ranking:} Use cosine similarity to generate the index of K1-NN of each image $
		L_1, L_2, ..., L_N$;\\
		\STATE {Neighborhood expansion}:\\
		\FOR {$j$=1,...,$N$}
		\STATE Initialize $num \leftarrow  0$;\\
		\FOR {$j$=1,...$N$}
		\STATE $num_j \leftarrow$ the size of ${L_i} \cap {L_j}$;\\
		\ENDFOR
		\STATE Sort $num$ by descending order and keep the top K2 $\{L_j\}$; \\
		\STATE Set new ${{L'}_i}$ $ \leftarrow $  union of the top K2 $\{L_j\}$;\\
		\ENDFOR
		\FOR {$j$=1,...,$N$}
		\STATE Construct \textbf{S} with new ${{L'}_i}$ base on Eq.\ref{eq.neigh.exp};\\
		\ENDFOR
		\RETURN $\textbf{S}$;\\
		\caption{Construct of neighborhood structure}
		\label{alg.pairwise}
	\end{algorithmic}
\end{algorithm}

The precision curve (evaluated using the labels) in Figure~\ref{fig1} indicates the quality of the constructed neighborhood for different values of $K1$. Due to rapidly decreasing precision with increasing $K1$, creating a large-enough neighborhood by simply increasing $K1$ is not the best option. In order to find a better approach, we borrow the ideas from the domain of graph modeling. In an undirected graph, if a node $v$ is connected to a node $u$ and if $u$ is connected to a node $w$, we can infer that $v$ is also connected to $w$. Inspired by this, if we treat every training image as a node in an undirected graph, we can expand the neighborhood of an image $i$ by exploring the neighbors of its neighbors. Specifically, if $\textbf{x}_{i}$ connects to $\textbf{x}_{j}$ and $\textbf{x}_{j}$ connects to $\textbf{x}_{k}$, we can infer that $\textbf{x}_{i}$ has the potential to be also connected to $\textbf{x}_{k}$.

In view of the above, we use the initial similarity matrix $\textbf{S}_1$ to expand the neighborhood structure. Specifically, based on $\textbf{S}_1$, we calculate the similarity of two images {by comparing the corresponding columns in $\textbf{S}_1$ using the expression }{$\frac{1}{||\left(\textbf{S}_1\right)_i-\left(\textbf{S}_1\right)_j||^2}$}. Then we again construct a ranked list of $K2$ neighbors, based on which we generate the second similarity matrix $\textbf{S}_2$ as:
\begin{eqnarray}
{{\left(\textbf{S}_2\right)}_{ij}} = \left\{ {\begin{array}{*{20}{l}}
	{1,~\textrm{if}~\textbf{x}_{j}~\textrm{is K2-NN of}~\textbf{x}_i}\\
	{-1,~\textrm{otherwise}}
	\end{array}} \right.
\end{eqnarray}
Finally, we construct the neighborhood structure by combining the direct and indirect similarities from the two matrices together. This results in the final similarity matrix $\textbf{S}$:
\begin{eqnarray}
{S_{ij}} = \left\{ {\begin{array}{*{20}{l}}
	{1,~\textrm{if}~{\left(\textbf{S}_1\right)}_{ij}=1~\textrm{or}~\textbf{x}_j~\textrm{is a K1-NN of}~\textbf{x}_i\textrm{'s~K2-NN}}\\
	{-1,~\textrm{otherwise}}
	\end{array}} \right.
\label{eq.neigh.exp}
\end{eqnarray}

The whole algorithm is shown in Alg.~\ref{alg.pairwise}. We note here that we could have also omitted this preprocessing step and construct the neighborhood structure directly during the learning of our neural network. We found, however, that the construction of neighborhood structure is time-consuming, and that updating of this structure based on the updating of image features in each epoch does not have significant impact on the performance. Therefore, we chose to obtain this neighborhood structure as described above and fix it for the rest of the process.

\subsection{Architecture Structure}
%\textcolor{blue}{Put the implementation details in the supplementary files.}
{As shown in Figure~\ref{fig.framework}}, our proposed BGAN consists of four components: encoder, hashing, generator and discriminator. We describe each of them in detail in the remainder of this section.

\eat{\begin{figure*}[t]
		\centering
		\includegraphics[width=0.9\linewidth]{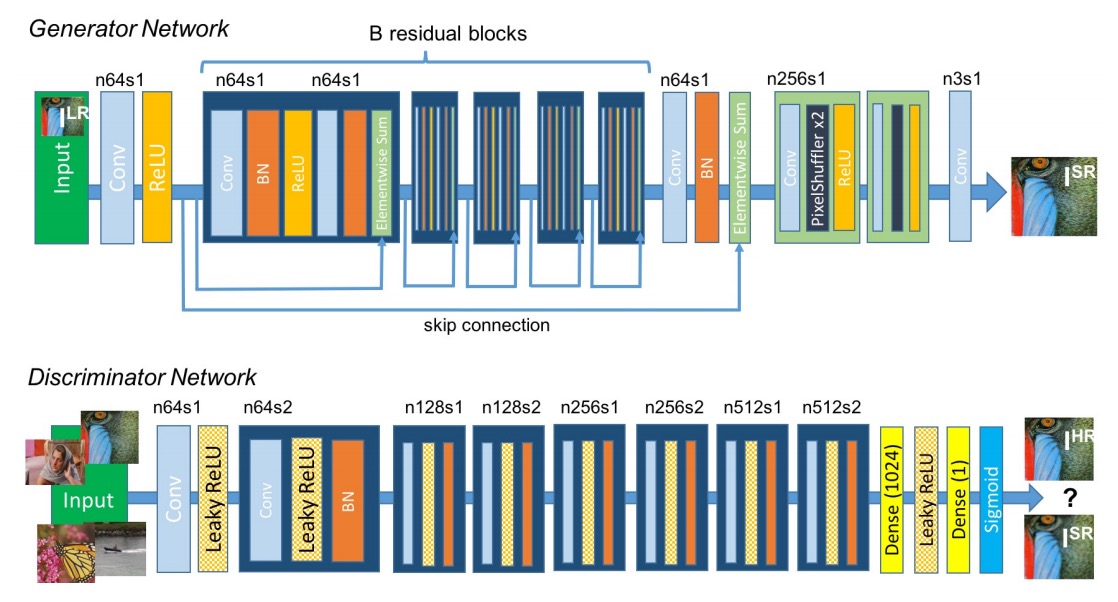}
		\caption{The architecture of BGAN. The generator is VAE. The descriminator is similar to this. (\textcolor{blue}{Replace this image})}
		\label{fig.architecture}
	\end{figure*}}
	
	\subsubsection{Encoder}
	%Our model is consisted of feature extraction, generator and discriminator, as shown in Fig.~\ref{fig.architecture}.
	For feature extraction, we use a structure similar to VGG19 \cite{GoogleNet}, with the details shown in table \ref{tab.arch}. We use 5 groups of convolution layers and 5 max convolution-pooling layers. Similar to \cite{GoogleNet}, we use 64, 128, 256, 512, 512 filters in the 5 groups of convolutional layers, respectively. %\textcolor[rgb]{1,0,0}{In the experiments, we also work with alternative feature extraction approaches, by replacing this module using deep networks, such as \cite{krizhevsky2012imagenet} and \cite{He2015}.}

	\begin{table}[]
		\centering
		\footnotesize
		\caption{The architecture for feature extraction}
		\label{tab.arch}
		\begin{tabular}{|c|c|c|c|}\hline
			Layer         & Size of Filter & Number of Filters & Others                  \\\hline\hline
			conv1\_1      & 3x3            & 64                & Stride=1,padding=1,relu \\
			conv1\_2      & 3x3            & 64                & Stride=1,padding=1,relu \\
			Max pooling   & 2x2            &                   & 2                       \\\hline
			conv2\_1      & 3x3            & 128               & Stride=1,padding=1,relu \\
			conv2\_2      & 3x3            & 128               & Stride=1,padding=1,relu \\
			Max pooling   & 2x2            &                   & 2                       \\\hline
			conv3\_1      & 3x3            & 256               & Stride=1,padding=1,relu \\
			conv3\_2      & 3x3            & 256               & Stride=1,padding=1,relu \\
			conv3\_3      & 3x3            & 256               & Stride=1,padding=1,relu \\
			Max pooling   & 2x2            &                   & 2                       \\\hline
			conv4\_1      & 3x3            & 512               & Stride=1,padding=1,relu \\
			conv4\_2      & 3x3            & 512               & Stride=1,padding=1,relu \\
			conv4\_3      & 3x3            & 512               & Stride=1,padding=1,relu \\
			Max pooling   & 2x2            &                   & 2                       \\\hline
			conv5\_1      & 3x3            & 512               & Stride=1,padding=1,relu \\
			conv5\_2      & 3x3            & 512               & Stride=1,padding=1,relu \\
			conv5\_3      & 3x3            & 512               & Stride=1,padding=1,relu \\
			Max pooling   & 2x2            &                   & 2                       \\\hline
			FC6           &                & 4096              & relu                    \\
			FC7           &                & 4096              & relu                    \\\hline
			%		Hashing Layer &                & L                 &                         \\
		\end{tabular}
	\end{table}
	
	\subsubsection{Hashing}
	
	A binary hash code is learned directly, by converting the $L$-dimensional representation $\textbf{z}$ learned from the last fully-connected layer FC7, which is continuous in nature, to a binary hash code $\textbf{b}$ taking values of either $+1$ or $-1$. This binarization process can only be performed by taking the sign function $\textbf{b} = sgn(\textbf{z})$ as the activation function on top of the hash layer.
	\begin{align*}
	\textbf{b} = {\mathop{\rm sgn}} (\textbf{z}) = \left\{ {\begin{array}{*{20}{l}}
		{ + 1,~if~\textbf{z} \ge 0}\\
		{ - 1,~~\textrm{otherwise}}
		\end{array}} \right.
	\label{eq.sgn}
	\end{align*}
	
	Unfortunately, as the sign function is non-smooth and non-convex, its gradient is zero for all nonzero inputs, and is ill-defined at zero, which makes the standard back-propagation infeasible for training deep networks. This is known as the	vanishing gradient problem, which has been a key difficulty in training deep neural networks via back-propagation [14]. Approximate solutions \cite{Zhang:2014:SHL:2600428.2609600,kang2016column} that relax the binary constraints are not a good alternative as they lead to a large quantization error and thereofre to a suboptimal solution \cite{Zhang:2014:SHL:2600428.2609600}.
	
	In order to tackle this challenge of optimizing deep networks with non-smooth sign activation, we draw inspiration from recent studies on continuation methods \cite{allgower2012numerical,cao2017hashnet}. These studies address a complex optimization problem by smoothing the original function, turning it into a different problem that is easier to optimize. By gradually reducing the amount of smoothing during the training, this results in a sequence of optimization problems converging to the original optimization problem. Following this approach, if we find an approximate smooth function of $sgn(.)$, and then gradually make the smoothed objective function non-smooth as the training proceeds, the final solution should converge to the desired optimization target.
	
	Motivated by the continuation methods, we define a function $app(.)$ to approximate $sgn(.)$:
	\begin{align}
	app(z) = \left\{ {\begin{array}{*{20}{l}}
		{ + 1,~if~\textbf{z} \ge 1}\\
		{~~~z,~~~~~if~1 \ge \textbf{z} \ge  - 1}\\
		{ - 1,~if~\textbf{z} \le  - 1}
		\end{array}} \right.
	\end{align}
	
\begin{figure}[t]
		\centering
		\subfigure[\small{tanh(.)}]{
			\label{fig.alg.tanh}
			\includegraphics[width=0.50\linewidth,height=3cm]{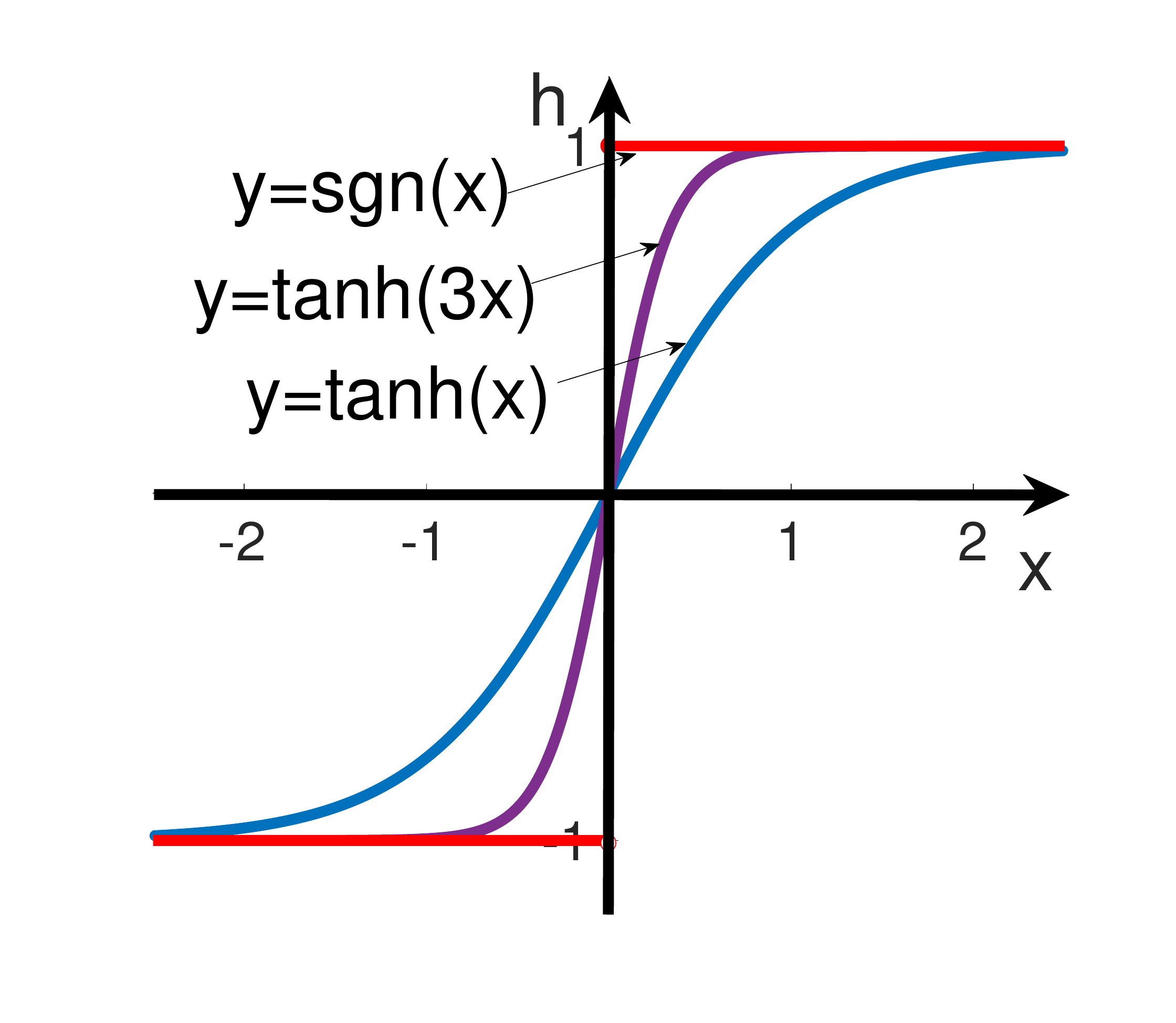}}
		\hspace{-0.5cm}
		\subfigure[\small{app(.)}]{
			\label{fig.alg.app}
			\includegraphics[width=0.50\linewidth,height=3cm]{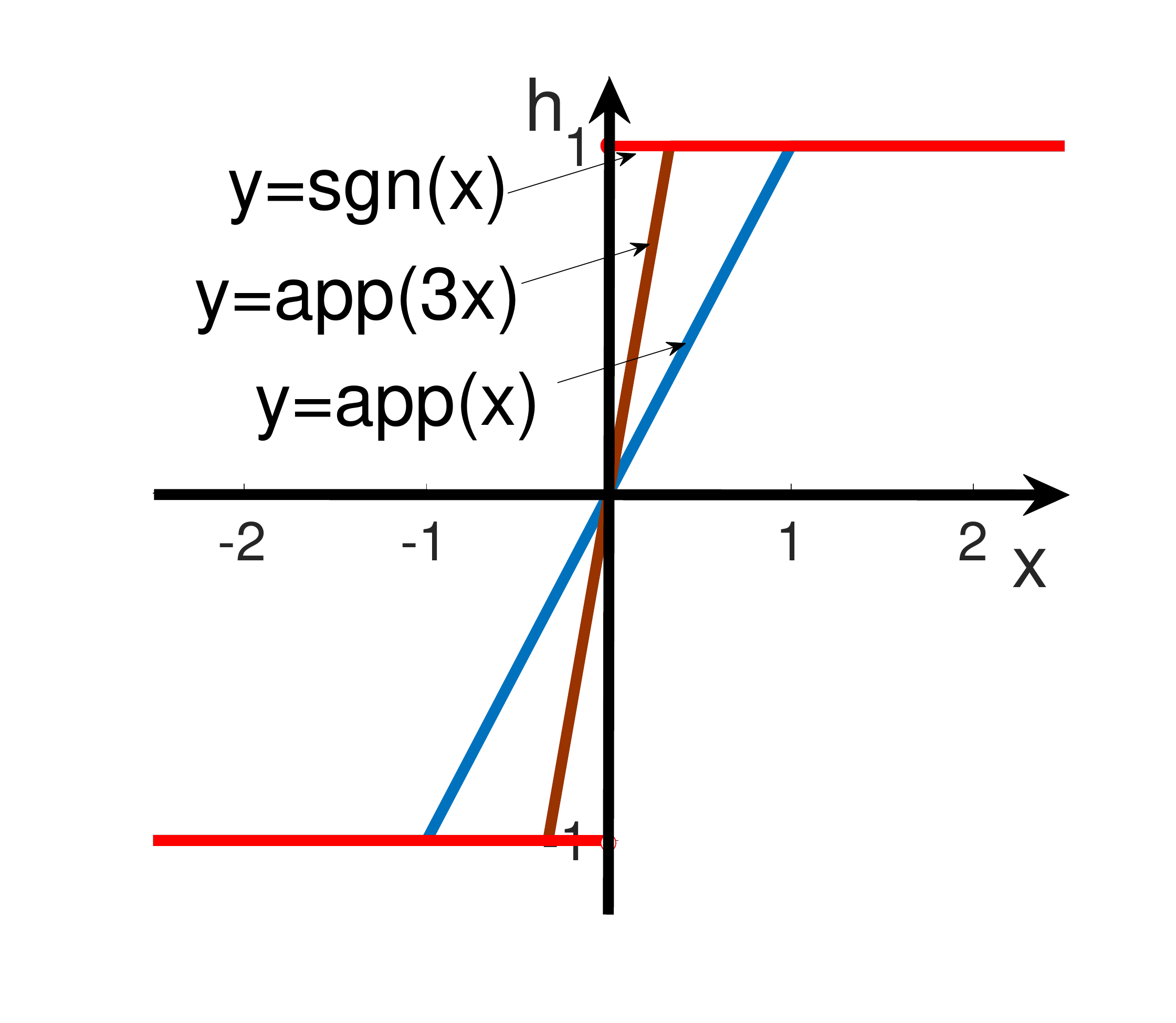}}
		\caption{An illustration of the process through which $app(.)$ approximates $sgn(.)$}
		\label{fig.alg.hash}
	\end{figure}

	We notice that there exists a key relationship between the sign function and the app function in the concept of limit:
	\begin{align}
	{\mathop{\rm sgn}} \left( \textbf{z} \right) = {\lim _{\beta  \to  + \infty }}app \left( {\beta \textbf{z}} \right)
	\label{eq.app}
	\end{align}
	An illustration of the process through which $app(.)$ approximates $sgn(.)$ is given in Fig.~\ref{fig.alg.hash}. The figure also shows the same process for an alternative to $app(.)$: $tanh(.)$:
	\begin{align}
	{\mathop{\rm sgn}} \left( \textbf{z} \right) = {\lim _{\beta  \to  + \infty }}\tan \left( {\beta \textbf{z}} \right)
	\label{eq.tanh}
	\end{align}

	\subsubsection{Generator and Discriminator}
	
	The task of our BGAN ``Generator'' network \textbf{G} is to generate an image based on the hash code \textbf{b}. 
	First, we let \textbf{b} serve as the input of the top fully-connected layer with the size $8\times8\times256$. Then, we use four deconvolutional layers with the size of kernels $5 \times 5, 5 \times 5, 5 \times 5, 1 \times 1$, and the number of kernels $256, 128, 32, 3$, which is followed by batch-normalization layers and eLU as the activation function.
	
	Following the approach by Goodfellow et al. \cite{goodfellow2014generative}, we define a ``Discriminator'' network \textbf{D} in such a way that it is optimized using criteria that are conflicting to those of \textbf{G}. In this way, \textbf{D} can act as adversary to \textbf{G} in the overall min-max optimization process. The goal of this optimization is to improve \textbf{G} such to be able to generate the images as well as possible. The process being adversarial to image generation is the process of trying to distinguish between the original and reconstructed images. If \textbf{G} manages to generate the images so well to ``fool'' \textbf{D}, then it ``wins'' the min-max game and the overall GAN optimization has converged. In view of this, given a model of the image classifier \textbf{D} assessing the original ($\textbf{I}$) and reconstructed ($\textbf{I}^R$) image, we can formally define the min-max game resulting in the optimal system parameters as follows: 
\begin{align}
\mathop {\min }\limits_{{\theta _\textbf{G}}} \mathop {\max }\limits_{{\theta _\textbf{D}}} \log \left( {\textbf{D}\left( \textbf{I} \right)} \right) + \log \left( {1 - \textbf{D}\left( {{\textbf{I}^R}} \right)} \right)
\end{align}

Here we follow the architecture design summarized by Radford et al. \cite{radford2015unsupervised}, use eLU activation and avoid max-pooling throughout the network. {It contains 4 convolutional layers with an increasing number of $5 \times 5$ filter kernels (32, 128, 256, and 512). Strided convolutions are used to reduce the image resolution each time the number of features is doubled. The resulting 512 feature maps are followed by a dense layer with the size of 1024 and a final sigmoid activation function to obtain a probability for sample classification.}

%We do the following modifications: Batch Normalization, LeakyReLU..... The Reason is that...

\subsection{Loss Function}

The definition of the loss function $\ell$ is critical for the performance of our network as it steers the overall optimization of the min-max game. While auto-encoding is commonly modeled based on the Mean Squared Error, we follow \cite{larsen2015autoencoding,LedigTHCATTWS16} and design a loss function that assesses our hashing solution with respect to perceptually relevant characteristics. We formulate the loss function as the weighted sum of a \textit{neighbor structure loss}, \textit{content loss} and \textit{adversarial loss} as:
\begin{eqnarray}
\ell = \ell_N + \lambda_1 \ell_C + \lambda_2 \ell_A
\label{eq.loss}
\end{eqnarray}

In the following subsections, we explain our realization of each of the loss functions.

\subsubsection{Neighbors Structure Loss}

The neighborhood structure loss models the loss in the similarity structure in data, as revealed in the set of neighbors obtained for an image by applying the hash code of that image. Given the binary codes $\textbf{B} = \left\{ {{\textbf{b}_i}} \right\}_{i = 1}^N$ for all the images of the length $L$ , and using the similarity matrix $\textbf{S}$ as the reference for similarity relations between images, we define the neighbors structure loss as follows:
\begin{equation}
\label{Eq.1}
\ell_N = \frac{1}{2}{\sum\limits_{{s_{ij}} \in S} {\left( {\frac{1}{L}{\textbf{b}_i}^T{\textbf{b}_j} - {s_{ij}}} \right)} ^2}
\end{equation}
The goal of optimizing for this loss function is clearly to bring the binary codes of similar images as close to each other as possible.

\subsubsection{Content Loss} 
The content loss models the loss in the quality of reconstructed images. The most widely used pixel-wise MSE loss is calculated as:
\begin{align}
{\ell_{MSE}} = \frac{1}{{WH}}\sum\limits_{i = 1}^W {\sum\limits_{j = 1}^H {{{\left( {{I_{ij}} - I_{ij}^R} \right)}^2}} } 
\label{eq.msr}
\end{align}
However, as pointed out in \cite{LedigTHCATTWS16}, solutions of MSE optimization problems often lack high-frequency content, which results in perceptually unsatisfying solutions with overly smooth textures. Also, element-wise reconstruction errors are not robust for images {and other signals with invariance}.

Therefore, instead of only relying on pixel-wise losses we build on the ideas of Christian et. al. \cite{LedigTHCATTWS16} and use a loss function that closer resembles perceptual similarity. 
Specifically, we define the VGG loss based on the eLU activation layers of the last convolutional layer of our discriminator networks \textbf{D}. With $\phi$ we indicate the feature map obtained by the last convolution (after activation) and then define the VGG loss as the Euclidean distance between the feature representations of a reconstructed image $I^R$ and the original image~$I$:
\begin{align}
{\ell_{Perceptual}} = \frac{1}{{WH}}\sum\limits_{i = 1}^W {\sum\limits_{j = 1}^H {{{\left( {\phi \left( {{\textbf{I}_{ij}}} \right) - \phi \left( {\textbf{I}_{ij}^R} \right)} \right)}^2}} }
\end{align}
Here, $W$ and $H$ represent the dimensions of the respective feature maps. Based on the above, we can now  define the content loss as:
\begin{align}
{\ell_C} = {\ell_{MSE}} + {\ell_{Perceptual}}.
\end{align}

\subsubsection{Adversarial Loss:} 
The adversarial loss models the loss due to misclassification of images, as done by \textbf{D}, in the original and reconstructed ones. Using the notations already explained in the context of Eq.8, we model this loss as:
\begin{align}
{\ell_A} = \log \left( {\textbf{D}\left( \textbf{I} \right)} \right) + \log \left( {1 - \textbf{D}\left( {{\textbf{I}^R}} \right)} \right)
\end{align}

%Please refer to Eq.6 in \cite{LedigTHCATTWS16}.

\subsection{{Learning}}
Using the loss function in Eq.\ref{eq.loss}, we train our network.
The forward propagation is as follows.
First, we use a deep convolutional network as the encoder to extract the features and then use hash layer to obtain to embed the real-valued features into binary codes:
\begin{eqnarray}{\textbf{b}_i} = sgn({\textbf{W}_h^T}\varphi  ({\textbf{I}_i};{\theta} ) ) \label{Eq.bm}\end{eqnarray}
where $\textbf{I}_i$ is an input image, ${\theta} $ is the parameter of the encoder and $\textbf{W}_h$ stands for the parameter of generating $\textbf{Z}$.
Then, $\textbf{b}_i$ is the input for a generator $\textbf{G}$ to reconstruct an image $\textbf{I}^R$:
\begin{eqnarray}
{\mathop{\rm \textbf{I}}\nolimits} _i^R = \phi ({\textbf{b}_i};\pi )
\end{eqnarray}
where $\pi$ stands for the parameters of the generator $\textbf{G}$.
Finally, a discriminator $D$ assigns probability:
\begin{eqnarray}
p = \textbf{D}(\textbf{I}^R_i;\psi )
\end{eqnarray}
that $\textbf{I}^R_i$ is an actual training sample and probability $1-p$ that $\textbf{I}^R_i$ is generated by our model ${\mathop{\rm \textbf{I}}\nolimits} _i^R = \phi ({\textbf{b}_i};\pi )$.

In the network, we have parameters of $\theta$, $\pi$, $\psi$ and $\textbf{W}_h$ to learn. We use back-propagation (BP) for learning and stochastic gradient descent (SGD) to minimize the loss.
In particular, we initialize network parameters and use forward propagation to obtain the value of each loss ($\ell_N,\ell_C,\ell_A$). In each iteration, we sample a mini-batch of images from the training set, and then update each parameter:
\begin{eqnarray}
\theta  \leftarrow  \theta - \tau\triangledown_\theta \left( {\ell_N + \ell_C} \right),\\
\pi  \leftarrow  \pi - \tau\triangledown_\pi \left( {\ell_C + \ell_A} \right),\\
\psi  \leftarrow  \psi + \tau\triangledown_\psi  { \ell_A} ,\\
\textbf{W}_h  \leftarrow  {\textbf{W}_h} - \tau\triangledown_{\textbf{W}_h} \left( {\ell_N + \ell_C} \right)
\end{eqnarray}
where $\tau$ is the learning rate. We train our network until it converges.

%Here we need to note that although our network is divided into three parts, we train the whole network in an end-to-end way and we use the raw images as the input and then update all the parameters.

For the hashing layer, we start training BGAN with $\beta_t =1$. For each stage $t$, after BGAN converges, we increase $\beta_t$ and train (i.e., fine-tune) BGAN by setting the converged network parameters as the initialization for training the BGAN in the next stage. By evolving $app(\beta \textbf{z})$ with $\beta_t \approx \infty$, the network will converge to BGAN with $sgn(\textbf{z})$ as activation function, which can generate exactly binary hash codes as we desire. Using $\beta_t = 10$ we can already achieve fast convergence for training BGAN.

%We optimized \ref{Eq.bm} by \textit{Neighbors Structure Loss} \ref{alg.pairwise}.

\eat{Specifically, reconstructing image from very short compact codes, which only have 16 or 48 bits, is very difficult for a decoder. Therefore, we project the short compact codes into a high dimension, here we use $16,384=8\times 8 \times 256$ to describe a compact information. To optimize the quality of the constructed image, we use $\textit{content loss}$ to minimize the decoder network. For the last part, the discriminator is to distinguish whether the input image is real or fake and output only one number $p$ which ranges from 0 to 1 and stands for the probability of the facticity of a input image. We can formulate this by:
\begin{eqnarray}
p = D(\widetilde{I}_i;\psi )
\end{eqnarray}
where  $\psi$ is the parameters of discriminator and $\widetilde{I}_i$ can be source input images or reconstructed images.
}

\section{Experiments}
\label{sec.exp}
We evaluate our BGAN on the task of large-scale image retrieval. 
Specifically, the experiments are designed to study the following research questions of our algorithm:
\\\noindent\textbf{RQ1}: How does each component of our algorithm affect the performance?
\\\noindent\textbf{RQ2}: Does the binary codes computed directly without relaxation improve the performance of the relaxed resolution?
\\\noindent\textbf{RQ3}: Does the performance of BGAN significantly outperform the state-of-the-art hashing algorithms?
\\\noindent\textbf{RQ4}: What is the efficiency of BGAN?

%Firstly, we study the influence of the parameters in our algorithm. Then, we compare our results with state-of-the-art algorithms on three standard datasets.
%\footnote{The code for SHPL will be released for public use.}

\subsection{Settings}
\subsubsection{Datasets}
We conduct empirical evaluation on three public benchmark datasets, CIFAR-10, NUS-WIDE, and Flickr.
\\\noindent\textbf{CIFAR-10} labeled subsets of the 80 million tiny images dataset, which consists of 60,000 $32\times 32$ color images in 10 classes, with 6,000 images per class. 
\\\noindent\textbf{NUS-WIDE} is a web image dataset containing 269,648 images downloaded from Flickr. Tagging ground-truth for 81 semantic concepts is provided for evaluation. We follow the settings in \cite{zhu2016deep} and use the subset of 195,834 images from the 21 most frequent concepts, where each concept consists of at least 5,000 images.
\\\noindent\textbf{Flickr} is a collection of 25,000 images from Flickr, where each image is labeled with one of the 38 concepts. We resize images of this subset into $256\times256$. 

%For CIFAR-10 and Flickr, similar to \cite{zhu2016deep}, we randomly select 5,000 training images and 1,000 test images in each class. For NUS-WIDE dataset, we randomly select 1,000 images as the test query set, and 4000 images as the training set.

In NUS-WIDE and CIFAR-10, we randomly select 100 images per class as the test query set, and 1,000 images per class as the training set. In Flickr, we randomly select 1,000 images as the test query set, and 4,000 images as the training set.

\subsubsection{Evaluation Metric}
Hamming ranking is used as the search protocol to evaluate our proposed approaches, and two indicators are reported.
1) Mean Average Precision (\textbf{mAP}): For a single query, Average Precision (AP) is the average of the precision value obtained for the set of top-k results, and this value is then averaged over all the queries.% The larger the MAP, the better the performance is.
2) \textbf{Precision}: We further use precision-recall curve and precision@K to evaluate the precision of retrieved images at each stage.

\eat{\begin{figure}[h]
 \centering
 \subfigure[\small{Performance variance with $N$}]{
  \label{fig.exp.N}
 \includegraphics[width=0.50\linewidth,height=3cm]{mapVSN-eps-converted-to.pdf}}
 \hspace{-0.5cm}
 \subfigure[\small{Number of iterations}]{
  \label{fig.exp.iter}
 \includegraphics[width=0.50\linewidth,height=3cm]{mapVSiter-eps-converted-to.pdf}}
 \caption{Parameters study with code length $32$ and $64$ on SIFT1M}
 \label{fig.exp.para}
 \end{figure}}

\subsubsection{Compared algorithms}
We compare our BGAN with other state-of-the-art hashing algorithms. Specifically, we compare with four non-deep hashing methods (iterative quantization (ITQ) hashing~\cite{GongLGP13}, spectral hashing (SH)~\cite{Weiss08}, Locality Sensitive Hashing (LSH)~\cite{Datar04} and spherical hashing~\cite{HeoLHCY15}), and two deep hashing methods (DeepBit~\cite{lin2016learning} and Deep Hashing (DH)~\cite{erin2015deep}).

To make a fair comparison, we also apply the non-deep hashing methods on deep features extracted by the CNN-F of the feature learning part in our BGAN. For non-deep hashing algorithms, we use 1134-D concatenated features as the hand-crafted features, including 64-D color histogram, 144-D color correlogram, 73-D edge direction histogram, 128-D wavelet texture, 225-D block-wise color moments and 500-D bag of words based on SIFT descriptions.

We also conduct the non-deep hashing methods on deep features extracted by the VGG network (VGG-fc7 \cite{GoogleNet}). 
For non-deep hashing algorithms, we use 512-dim GIST features as the hand-crafted features. We also compare with PCAH~\cite{WangKC12}, DGH~\cite{LiuMKC14}, AGH~\cite{LiuWKC11}, and UN-BDNH~\cite{do2016learning} in term of precision.

By constructing the neighborhood structure using the labels, our method can be easily modified as a supervised hashing method. Therefore, we also compare with some supervised hashing methods, e.g., iterative quantization hashing (ITQ-CCA)~\cite{GongLGP13}, KSH~\cite{LiuWJJC12}, minimal loss hashing (MLH)~\cite{NorouziF11}, DNNH~\cite{LaiPLY15}, CNNH \cite{CNNH} and Deep Hashing Network (DHN)~\cite{zhu2016deep}.

%To guarantee that our results directly comparable to most published results, the results of LSH, ITQ, MLH and SH are directly reported from the latest work (\cite{zhu2016deep}), while missing results are obtained by the implementations provided by their authors.

%For fair comparison, all of the methods use identical training and test sets.
\subsubsection{Implementation Details}
In the construction of neighborhood structure step, we set $K1=20,K2=30$ for CIFAR-10 dataset, and the average number of the neighbors for each image is $400,1021,1168$ for the three datasets. %If the number of neighbors is too big, the accuracy is not guaranteed. On the other hand, if we use too small number of neighbors, we are unable to fully explore the neighborhood structure and the performance of hash codes will decrease. Therefore, we empirically set the $K1,K2$ to the above values.
By default, we set $\lambda_1=0.1$ and $\lambda_2=0.1$. We set the mini-batch size as 256, and the learning rate as $0.01$.

\subsection{Component Analysis (RQ1)}
\begin{table}[b]
	\centering
	\caption{The mAP of BGAN on CIFAR-10 using different combinations of components.}
	\label{tab.comp}
	\begin{tabular}{l|c|c|c}\hline
		& \multicolumn{3}{c}{mAP}           \\\hline
		Components &  24-bit & 32-bit & 48-bit \\\hline
		$\ell_N$         & 0.487  &    0.511  &  0.543      \\
		$\ell_C$         &  -     &   -    &   -    \\
		$\ell_A$         &   -     &   -   &    -   \\
		$\ell_N+\ell_C$      &   0.247     &  0.379      &   0.497     \\
		$\ell_N+\ell_A$      &   0.472     &  0.503      &   0.534     \\
		$\ell_C+\ell_A$      &   -    &   -    &   -    \\\hline
		$\ell_N+\ell_C+\ell_A$  &\textbf{0.512} &  \textbf{0.531} &  \textbf{0.558} \\\hline
	\end{tabular}
\end{table}

\begin{table*}[t]
	\centering
	\small
	\caption{mAP for different unsupervised hashing methods using different number of bits on three image datasets}
	\label{tab.res.map.comp}
	\begin{tabular}{l|c|c|c|c|c|c|c|c|c|c|c|c}\hline
		\multirow{2}{*}{Method} & \multicolumn{4}{c|}{CIFAR-10}          & \multicolumn{4}{c|}{NUS-WIDE}          & \multicolumn{4}{c}{Flickr}            \\\cline{2-13}
		& 12 bits & 24 bits & 32 bits & 48 bits & 12 bits & 24 bits & 32 bits & 48 bits & 12 bits & 24 bits & 32 bits & 48 bits \\\hline\hline
		ITQ~\cite{GongLGP13}            & 0.162   & 0.169   & 0.172   & 0.175   & 0.452   & 0.468   & 0.472   & 0.477   & 0.544   & 0.555   & 0.560   & 0.570   \\
		SH \cite{Weiss08}               & 0.131   & 0.135   & 0.133   & 0.130   & 0.433   & 0.426   & 0.426   & 0.423   & 0.531   & 0.533   & 0.531   & 0.529   \\
		%KMH \cite{HeWS13}               & 0.134   & 0.137   & 0.139   & 0.142   & 0.***   & 0.***   & 0.***   & 0.***   & 0.***   & 0.***   & 0.***   & 0.***   \\
		LSH \cite{Datar04}              & 0.121   & 0.126   & 0.120   & 0.120   & 0.403   & 0.421   & 0.426   & 0.441   & 0.499   & 0.513   & 0.521   & 0.548   \\
		Spherical \cite{HeoLHCY15})     & 0.138   & 0.141   & 0.146   & 0.150   & 0.413   & 0.413   & 0.424   & 0.431   & \textbf{0.569}   & \textbf{0.559}   & \textbf{0.583}   & \textbf{0.572}   \\\hline\hline
		ITQ+VGG                 		& 0.196   & 0.246   & \textbf{0.289}   & \textbf{0.301}   & 0.435   & 0.435   & 0.548   & 0.435   & 0.553   & 0.548   & 0.545   & 0.560   \\
		SH+VGG                  		& 0.174   & 0.205   & 0.220   & 0.232   & 0.433   & 0.426   & 0.426   & 0.423   & 0.550   & 0.544   & 0.541   & 0.545   \\
		%KMH+VGG                 		&         &         &         &         &         &         &         &         &         &         &         &         \\
		LSH+VGG                 		& 0.101   & 0.128   & 0.132   & 0.169   & 0.401   & 0.442   & 0.480   & 0.471   & 0.543   & 0.549   & 0.555   & 0.551   \\
		Spherical+VGG           		& \textbf{0.212}   & \textbf{0.247}   & 0.256   & 0.281   & \textbf{0.549}   & \textbf{0.614}  & \textbf{0.653}   & \textbf{0.678}   & 0.552   &  0.547  & 0.546   & 0.545   \\\hline\hline
		DeepBit \cite{lin2016learning}  & 0.185   & 0.218   & 0.248   & 0.263   & 0.383   & 0.401   & 0.403   & 0.412   & 0.501   &  0.505  & 0.511   & 0.513   \\
		DH \cite{erin2015deep}			& 0.160   & 0.164   & 0.166   & 0.168   & 0.422   & 0.448   & 0.480   & 0.493   & 0.553   &  0.548  & 0.543   & 0.556   \\
		BGAN    						& \textbf{0.401}   & \textbf{0.512}   & \textbf{0.531}   & \textbf{0.558}   & \textbf{0.675}   &  \textbf{0.690}  & \textbf{0.714}   &  \textbf{0.728}  & \textbf{0.683}   & \textbf{0.702}   & \textbf{0.703}   & \textbf{0.703}     \\\hline
	\end{tabular}
\end{table*}

Our loss function consists of three major components: neighborhood structure loss ($\ell_N$), content loss ($\ell_C$) and adversarial loss ($\ell_A$). In this subsection, we study the effect of each component on the performance. Due to the space limit, we only report the results on the CIFAR-10 dataset in Table ~\ref{tab.comp}.
An interesting observation is that if we do not use the neighborhood structure loss ($\ell_N$), we will learn identical hash codes for all the images. A possible reason is that when we define no loss function directly on the hash codes, the loss from $\ell_C$ and $\ell_A$ in the back-propagation step cannot well guide the learning of hashing codes.
Another interesting finding is that using the combination of $\ell_N$ and $\ell_C$, or the combination of $\ell_N$ and $\ell_A$ can obtain even worse performance than using $\ell_N$ only.
The is understandable, because if we use $\ell_N$ and $\ell_C$ without $\ell_A$, the discriminator network $\textbf{D}$ based on $\ell_A$ cannot learn satisfactory features. Thus, the loss function $\ell_C$ based on the features of $\textbf{D}$ will mislead the learning of hash codes. It is the same case for the unsatisfactory performance of $\ell_N$ and $\ell_A$. When we only force the reconstructed images to be similar to the general images, without considering its similarity to the corresponding input images, the performance of the learned hash codes will also be degraded.

The best performance is achieved when we use the combination of the three components: $\ell_N + \ell_C + \ell_A$. Compared with using $\ell_N$ only, the performance is improved by 2.5\%, 2\% and 1.3\% for 24, 32, and 48-bit hash codes, which is not that significant.
%It is supposed to be further improved after we tune the parameters of $\lambda_1$ and $\lambda_2$.
From the above analysis, we can conclude that all these three components contribute to the great performance of our BGAN.

\subsection{Effect of Binary Optimization (RQ2)}
As discussed above, BGAN can learn binary hash codes directly while previous hashing methods first learn continuous representations and then generate hash codes using a sign function as post-process. In this subsection, we study the effect of direct binary codes optimization on the performance of hash codes, and the results are shown in Table~\ref{tab.binary}.
As shown in Table~\ref{tab.binary}, our binary optimization can improve the performance of the learned binary codes.
Specifically, the first solution (Eq.\ref{eq.app}) outperforms two-step solution by 2.5\%, 3.2\%, and 2.3\% for 24, 32, and 48-bit hash codes. While the second solution (Eq.\ref{eq.tanh}) improves it by 3.0\%, 4.6\%, and 2.8\%.
This verifies our argument that two-step solution is sub-optimal, and binary optimization can achieve a better performance.
\begin{table}[h]
	\centering
	\caption{The mAP of BGAN on CIFAR-10 using different combinations of components.}
	\label{tab.binary}
	\begin{tabular}{l|c|c|c}\hline
		& \multicolumn{3}{c}{mAP}           \\\hline
		Methods & 24-bit & 32-bit & 48-bit \\\hline
		two-step solution          & 0.512       & 0.531       & 0.558       \\
		${\mathop{\rm sgn}} \left( z \right) \!=\! {\lim _{\beta  \to  + \infty }}\tanh \left( {\beta z} \right)$               &   \textbf{0.542}     & \textbf{0.577}       &  \textbf{0.586}     \\
		${\mathop{\rm sgn}} \left( z \right) \!=\! {\lim _{\beta  \to  + \infty }}app \left( {\beta z} \right)$               &   0.537     & 0.563       &    0.581       \\\hline
	\end{tabular}
\end{table}

\begin{figure*}[t]
	\centering
	\subfigure[16 bits]{
		\includegraphics[width=0.30\linewidth,height=3.6cm]{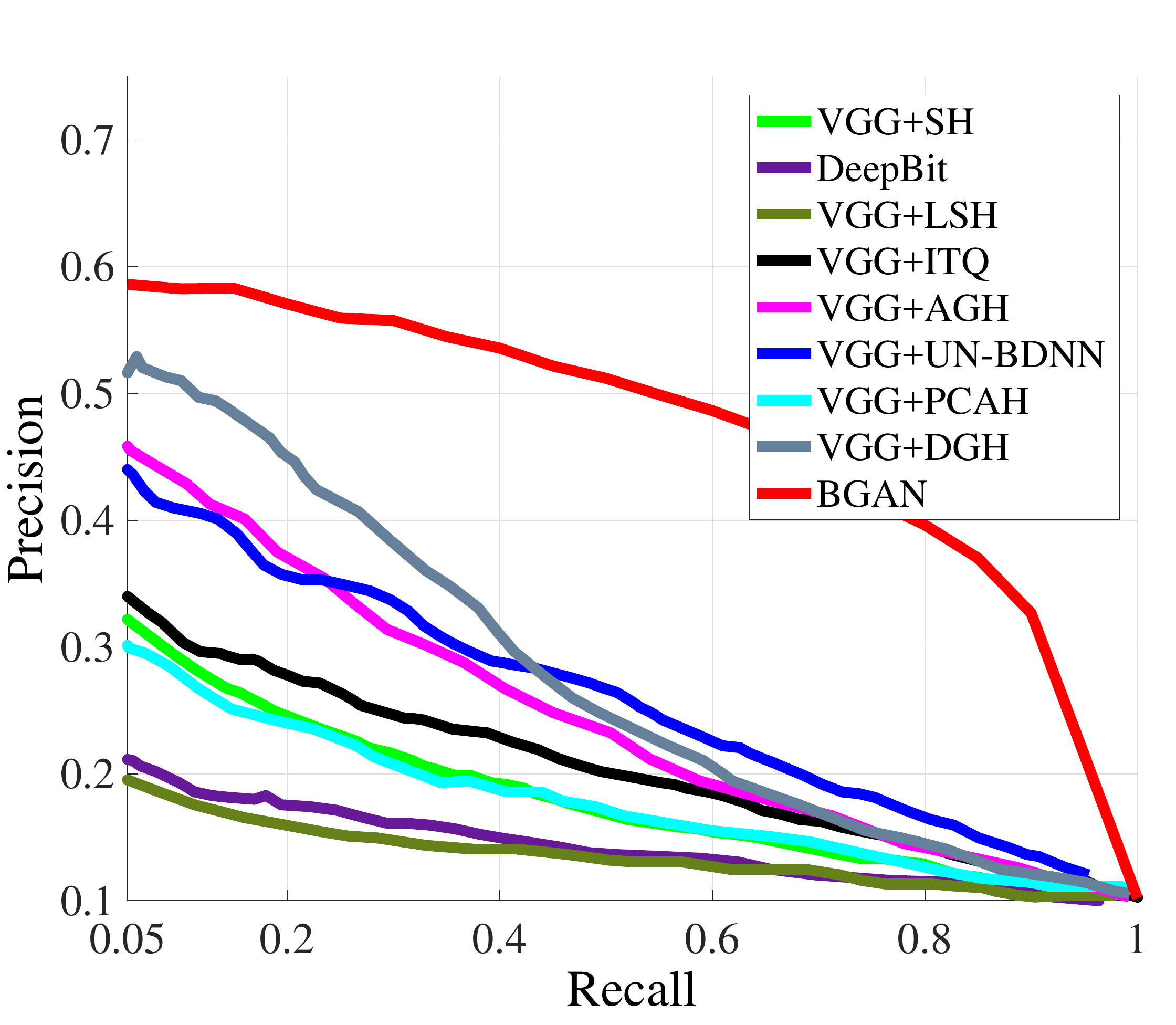}}
	\vspace{-0.15cm}
	\subfigure[32 bits]{
		\includegraphics[width=0.30\linewidth,height=3.6cm]{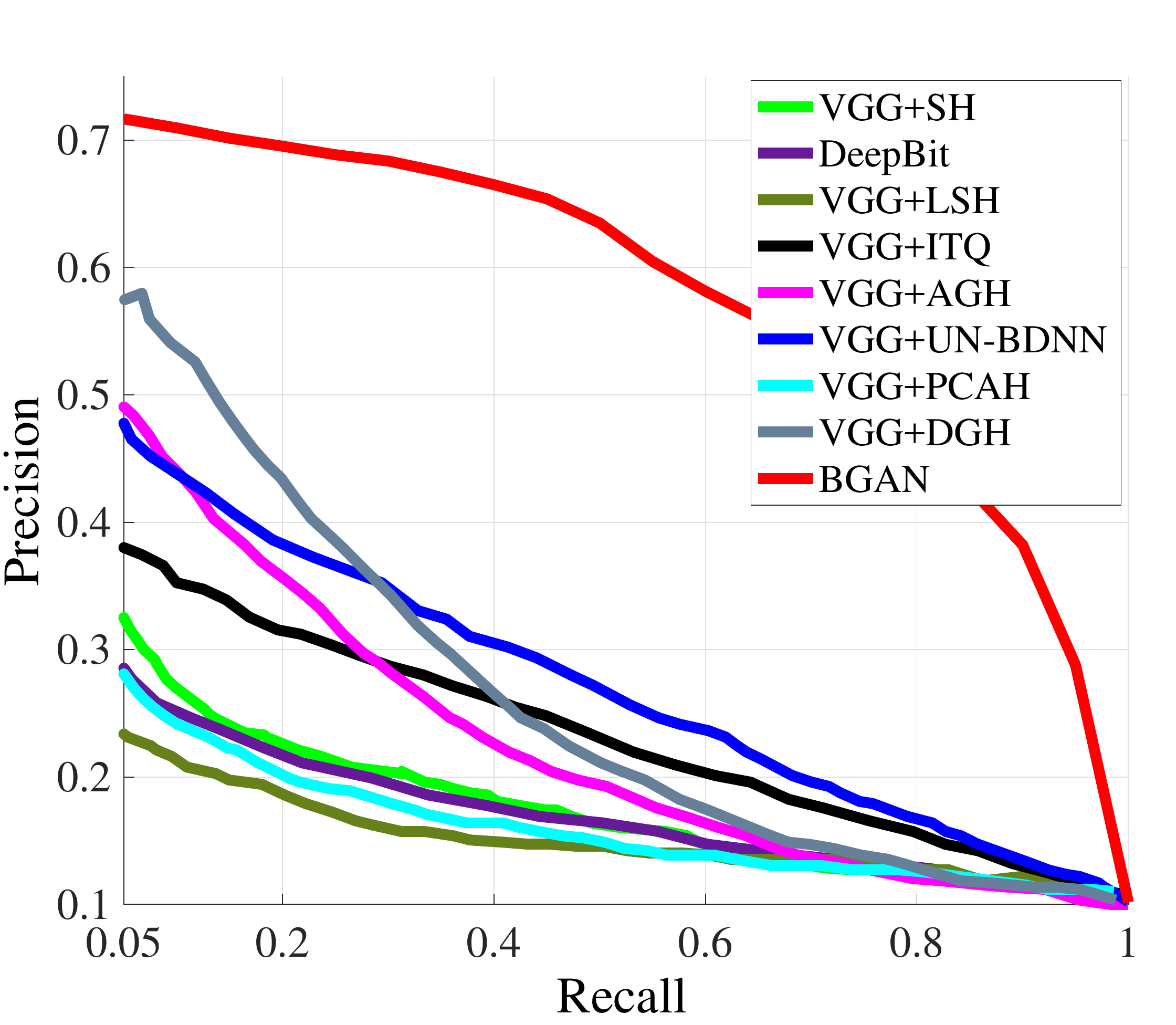}}
	\vspace{-0.15cm}
	\subfigure[64 bits]{
		\includegraphics[width=0.30\linewidth,height=3.6cm]{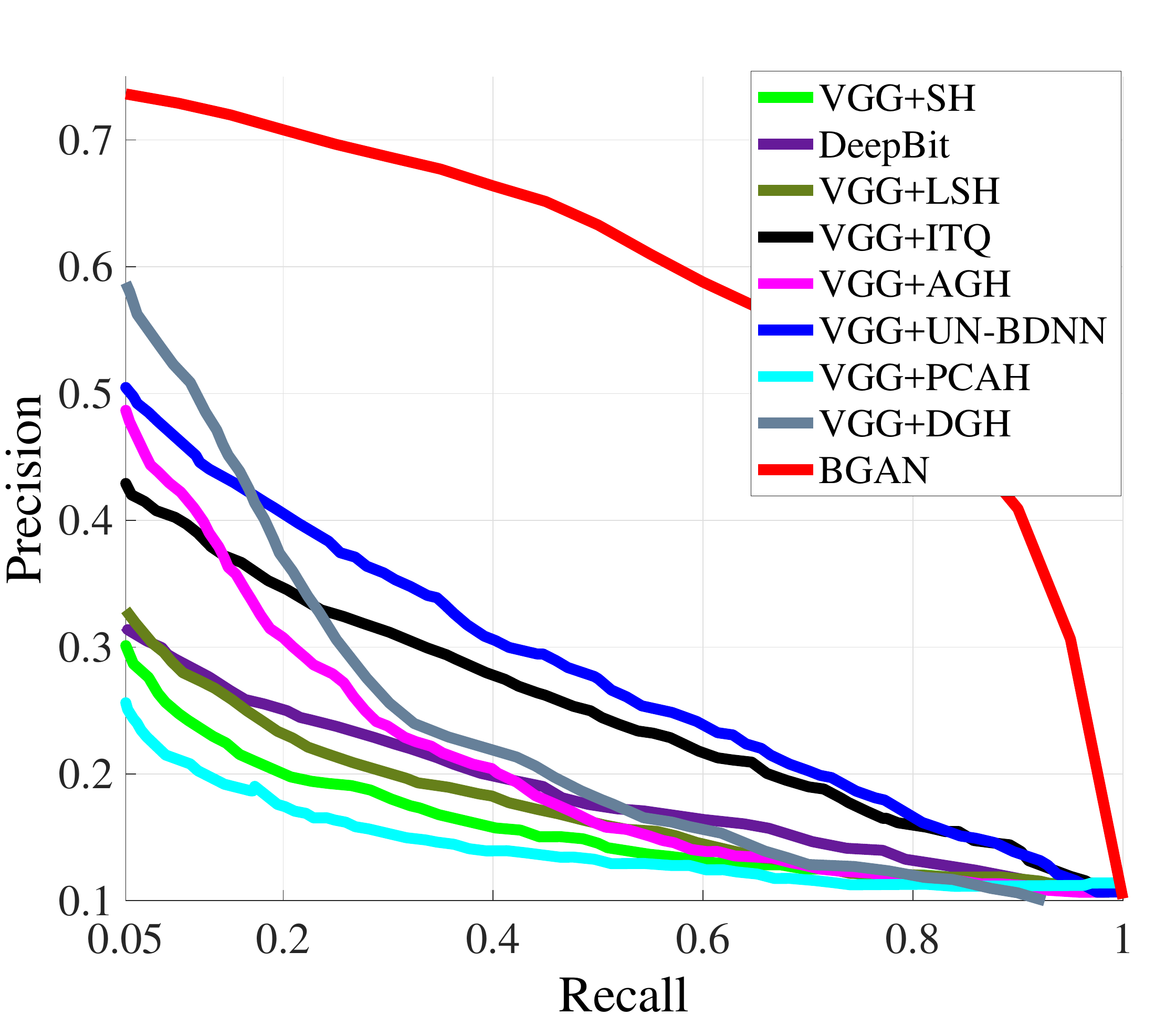}}
	\vspace{-0.15cm}
	\subfigure[16 bits]{
		\includegraphics[width=0.30\linewidth,height=3.6cm]{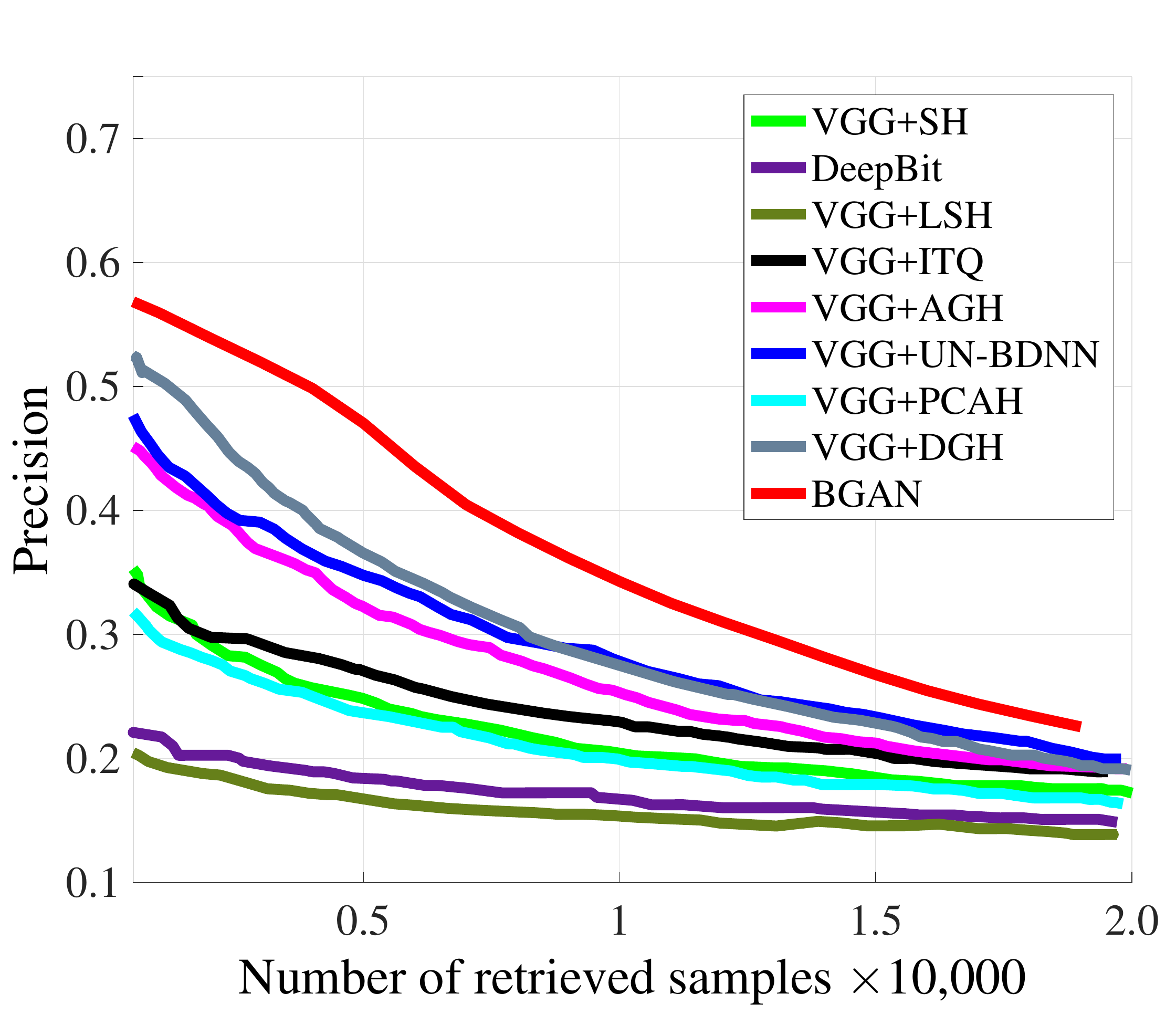}}
		\vspace{-0.15cm}
	\subfigure[32 bits]{
		\includegraphics[width=0.30\linewidth,height=3.6cm]{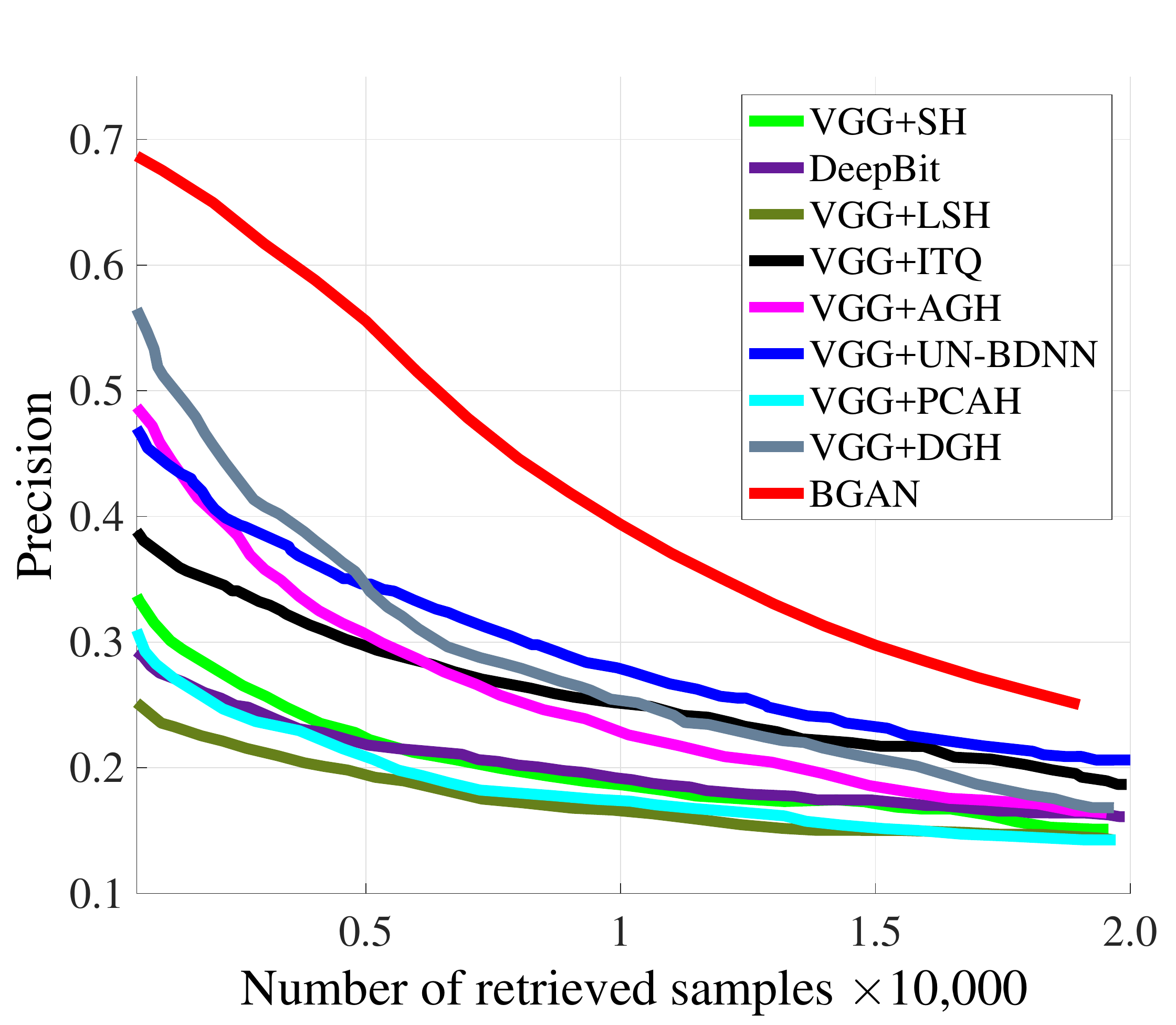}}
		\vspace{-0.15cm}
	\subfigure[64 bits]{
		\includegraphics[width=0.30\linewidth,height=3.6cm]{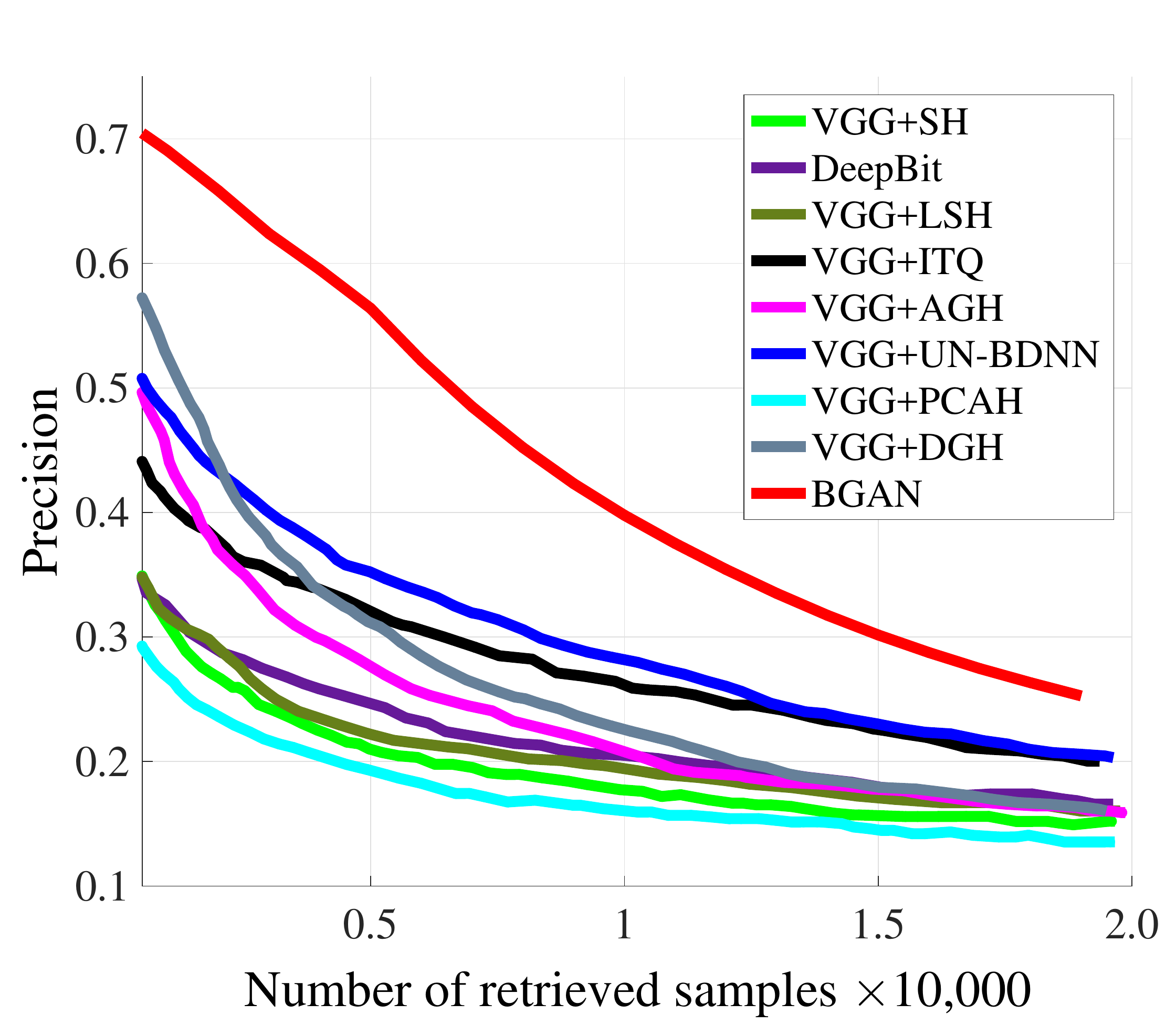}}
		\vspace{-0.15cm}
	\caption{Precision for different unsupervised hashing methods using different number of bits on CIFAR-10 dataset.}
	\label{fig.pr}
\end{figure*}

\begin{table}[t]
	\centering
	\caption{mAP for different supervised hashing methods using different number of bits on three image datasets}
	\label{tab.res.time}
	\begin{tabular}{c|c|c|c|c}\hline
		\multirow{2}{*}{Method} & \multicolumn{4}{c}{CIFAR-10}                \\\cline{2-5}
		& 12 bits & 24 bits & 32 bits & 48 bits  \\\hline\hline
		ITQ-CCA~\cite{GongLGP13}                 & 0.435   & 0.435   & 0.435   & 0.435     \\
		KSH~\cite{LiuWJJC12}                     & 0.556   & 0.572   & 0.581   & 0.588          \\
		MLH~\cite{NorouziF11}                    & 0.500   & 0.514   & 0.520   & 0.522    \\
		DNNH~\cite{LaiPLY15}                     & 0.674   & 0.697   & 0.713   & 0.715   \\
		CNNH\cite{CNNH}                          & 0.611   & 0.618   & 0.625   & 0.608      \\
		DHN~\cite{zhu2016deep}                   & \textbf{0.708}   & \textbf{0.735}   & \textbf{0.748}   & \textbf{0.758}         \\\hline
		BGAN                 	& \textbf{0.866}   & \textbf{0.874}   & \textbf{0.876}   & \textbf{0.877}    \\\hline
	\end{tabular}
\end{table}
\subsection{Compare with the State-of-the-art Algorithms (RQ3)}

\eat{The results are copied from these papers:\\
%$http://www.cv-foundation.org/openaccess/content_cvpr_2015/papers/Liong_Deep_Hashing_for_2015_CVPR_paper.pdf$,
Deep Hashing Network for Efficient Similarity Retrieval\\
Deep Hashing for Compact Binary Codes Learning\\
Deep Supervised Hashing for Fast Image Retrieval}

In this subsection, we compare our BGAN with different unsupervised hashing methods on three datasets. The results on mAP are shown in Table~\ref{tab.res.map.comp}, and the precision is shown in Fig.~\ref{fig.pr}. %We have tested $L=12, 24, 32, 48$. 
From Table~\ref{tab.res.map.comp} and Fig.~\ref{fig.pr}, we have the following observations:

\noindent 1) Our method significantly outperforms the other deep or non-deep hashing methods in all datasets. In CIFAR-10, the improvement of BGAN over the other methods is more significant, compared with that in NUS-WIDE and Flickr datasets. Specifically, it outperforms the best counterpart (Spherical+VGG) by 18.9\%, 26.5\%, 27.5\% and 27.7\% for 12, 24, 32 and 48-bit hash codes. One possible reason is that CIFAR-10 contains simple images, and the constructed neighborhood structure is more accurate than the other two datasets. BGAN improves the state-of-the-arts by 12.6\%, 7.6\%, 6.1\% and 5.0\% in NUS-WIDE dataset, and 11.4\%, 14.3\%, 12.0\% and 13.1\% in NUS-WIDE dataset.
%Also, the improvements gap between BGAN and other hashing methods is larger in the case of $L=12$, than that of $L=24,32,48$.
\eat{\item CKMeans achieves the best performance in terms of recall compared with all the other hashing methods, and the improvements over state-of-the-art hashing methods is $5\%$ to $20 \%$. However, it requires more time to perform the search, as is shown in Table~\ref{table.timeCost}. More than 10s is required to perform a 1000-NN search on the SIFT1B 32-bit dataset, which is around 100 times slower than SHPL.}

\noindent 2) Table~\ref{tab.res.map.comp} shows that Spherical+VGG is a strong competitor in terms of mAP, and Fig.~\ref{fig.pr} performs well for small number of retrieved samples (or recall). On the other hand, the performance of deep hashing methods (DeepBit~\cite{lin2016learning} and DH~\cite{erin2015deep}) is not superior. A possible reason is the deep hashing methods use only 3 full connected layers to extract the features, which is not very powerful.

\noindent 3) When we run the non-deep hashing methods on deep features, the performance is usually improved compared with the hand-crafted features. The performance gap is larger in CIFAR-10 and NUS-WIDE datasets than in Flickr dataset.

\noindent 4) With the increase of code length, the performance of most hashing methods is improved accordingly. More specifically, the mAP improvements using deep features are generally more significant than that of non-deep features in CIFAR-10 dataset and NUS-WIDE dataset. An exception is SH, which has no improvement with the increase of code length.

We also compared with supervised hashing methods, and shown the mAP results on CIFAR-10 dataset in Table~\ref{tab.res.map.comp.su}. It is obvious that our BGAN\_s outperforms the state-of-the-art deep and non-deep supervised hashing algorithms by a large margin, which are 15.8\%, 13.9\%, 12.8\% and 11.9\% for 12, 24, 32, and 48-bits hash codes. This indicates that the performance improvement of BGAN is not only due to the constructed neighborhood structure, but also the other components.

\vspace{-0.3cm}
\subsection{The Study of Efficiency (RQ4)}
\begin{figure}[t]
	\centering
	\includegraphics[width=0.6\linewidth,height=3cm]{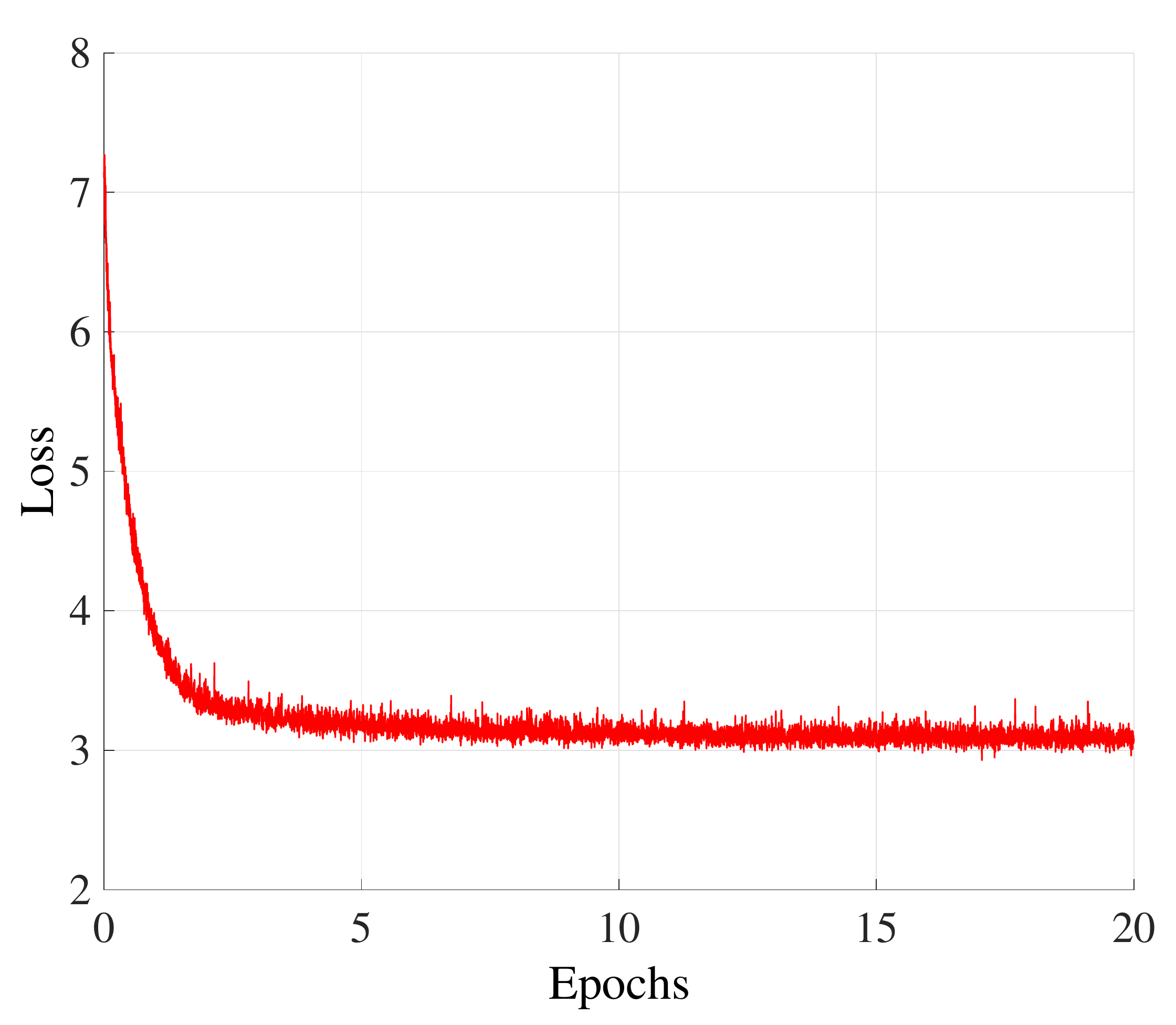}
	\caption{The convergence of BGAN in CIFAR-10 dataset.}
	\label{fig.conv}
\end{figure}

In this subsection, we study the efficiency of our algorithm. First, we study the convergence of our BGAN in CIFAR-10 dataset, and the results on shown in Fig.~\ref{fig.conv}.
It can be seen that our method converges after a few epochs, which shows the efficiency of our solution.

We also report the training and testing time in CIFAR-10 dataset of our algorithm, and compare it with DH \cite{erin2015deep} in Table \ref{tab.res.map.comp.su}. Since our BGAN has more parameters, it takes longer time for training and testing. However, BGAN is still very fast for generating the hash codes for a test image.

\begin{table}[h]
	\centering
	\caption{The training and testing time for BGAN and DH.}
	\label{tab.res.map.comp.su}
	\begin{tabular}{c|c|c}\hline
		Methods & Training time & Testing time   \\\hline\hline
% 		DeepBit \cite{lin2016learning}  & 0.   & 0.     \\
		DH \cite{erin2015deep}			& 1 hour   & 0.5 ms    \\
		BGAN                 	& 5h   & 3 ms    \\\hline
	\end{tabular}
\end{table}

\vspace{-0.3cm}
\subsection{Reconstruction Results}
\begin{figure}[t]
	\centering
	\includegraphics[width=1\linewidth]{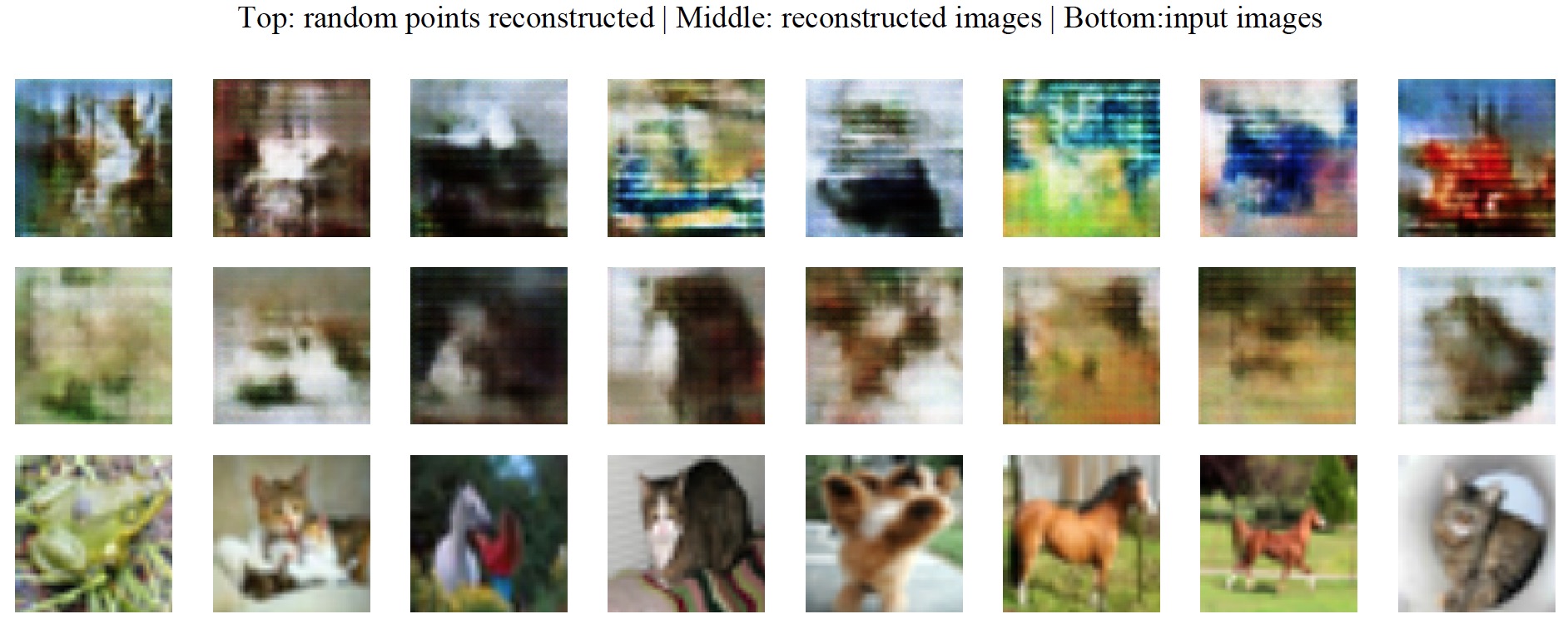}
	\caption{Image reconstruction using binary codes.}
	\label{fig.recons}
\end{figure}

To evaluate the ability of image reconstruction using BGAN, we show some qualitative results on CIFAR-10 dataset in Fig.~\ref{fig.recons}. The top row are the reconstructed results using random input, the second row are the reconstructed results from BGAN, and the last row are the original images. We can see that BGAN can reconstruct images which are similar to the original images.

\vspace{-0.3cm}
\section{Conclusion}
\label{sec.conclusion}
In this work, we propose an unsupervised hashing which addresses two central problems remaining largely unsolved for image hashing: 1) how to directly generate binary codes without relaxation? 2) how to equip the binary representation with the ability of accurate image retrieval, beyond vivid image generation? 
First, we propose two equivalent but smoothed activation functions, and design a learning strategy whose solution converges to the results of sign activation.
Second, we propose a loss function which consists of an adversarial loss, a content loss, and a neighborhood structure loss. 
%The adversarial loss pushes the generated image to the natural image manifold using a discriminator network that is trained to differentiate between the generated images and original images. The content loss enforces the generated image to be aligned with the conditioned original image, and a neighborhood structure loss is to exploit the data distributions of the images feature space.
Experimental results show that our BGAN doubled the performance of the state-of-the-arts in CIFAR-10 dataset, and also significantly outperformed the others on NUSWIDE, and Flickr datasets. In the future, it is necessary to improve the reconstruction accuracy of BGAN.


\end{document}